\documentclass[final]{cvpr}

\usepackage{times}
\usepackage{epsfig}
\usepackage{graphicx}
\usepackage{amsmath}
\usepackage{amssymb}
\usepackage{dblfloatfix}
\usepackage{wrapfig}
\usepackage{subfigure}

\usepackage[pagebackref=true,breaklinks=true,colorlinks,bookmarks=false]{hyperref}

\def\cvprPaperID{7448} 
\def\confYear{CVPR 2021}

\begin{document}

\title{TransNAS-Bench-101: Improving Transferability and Generalizability of Cross-Task Neural Architecture Search}

\author{Yawen Duan$^{1,}$\thanks{Equal contribution. (\{kmdaniel, cyn0531\}@connect.hku.hk)}\ , 
Xin Chen$^{1,*}$, 
Hang Xu$^{2}$, 
Zewei Chen$^{2}$, 
Xiaodan Liang$^{3, }$\thanks{Corresponding author. (xdliang328@gmail.com)}\ , 
Tong Zhang$^{4}$, 
Zhenguo Li$^{2}$\\
$^1$ The University of Hong Kong, 
$^2$ Huawei Noah's Ark Lab, 
$^3$ Sun Yat-sen University, \\
$^4$ The Hong Kong University of Science and Technology
}

\maketitle

\begin{abstract}

Recent breakthroughs of Neural Architecture Search (NAS) extend the field's research scope towards a broader range of vision tasks and more diversified search spaces. While existing NAS methods mostly design architectures on a single task, algorithms that look beyond single-task search are surging to pursue a more efficient and universal solution across various tasks. 
Many of them leverage transfer learning and seek to preserve, reuse, and refine network design knowledge to achieve higher efficiency in future tasks. 
However, the enormous computational cost and experiment complexity of cross-task NAS are imposing barriers for valuable research in this direction. 
Existing NAS benchmarks all focus on one type of vision task, i.e., classification. 
In this work, we propose TransNAS-Bench-101, a benchmark dataset containing network performance across seven tasks, covering classification, regression, pixel-level prediction, and self-supervised tasks. This diversity provides opportunities to transfer NAS methods among tasks and allows for more complex transfer schemes to evolve. We explore two fundamentally different types of search space: cell-level search space and macro-level search space. With 7,352 backbones evaluated on seven tasks, 51,464 trained models with detailed training information are provided. With TransNAS-Bench-101, we hope to encourage the advent of exceptional NAS algorithms that raise cross-task search efficiency and generalizability to the next level. 
Our dataset file will be available at Mindspore\footnote{https://download.mindspore.cn/dataset/TransNAS-Bench-101}, VEGA\footnote{https://www.noahlab.com.hk/opensource/vega/page/doc.html?\\path=datasets/transnasbench101}.

\end{abstract}

\vspace{-1mm}

\section{Introduction}

In recent years, networks found by Neural Architecture Search (NAS) methods are surpassing human-designed ones, setting state-of-the-art performance on various tasks \cite{tan2019efficientnet, real2019regularized}. Ultimately, NAS calls for algorithmic solutions that can discover near-optimal models for any task within any search space under moderate computational budgets. To pursue this goal, recent research in NAS quickly expanded its scope into broader vision domains such as object detection \cite{xu2019auto} and semantic segmentation \cite{chen2018searching}. \cite{shi2020bridging} seeks to bridge the gap between sample-based and one-shot approaches. Besides searching for an optimal cell design in earlier works \cite{liu2018darts}, many recent works also investigate the macro skeleton search of a network \cite{yao2019sm, xu2019auto}. 

Although many algorithms successfully shortened the search time of NAS from months to hours \cite{liu2018darts, dong2019searching}, some research has shown their reliance on specific search spaces and datasets \cite{yang2019evaluation}. There are also questions on these algorithms' efficiency when dealing with a large number of tasks \cite{chen2020catch}. A rising direction of NAS research is thus looking for solutions that acquire transferable knowledge and generalize well across multiple tasks and search spaces \cite{pasunuru2019continual, wong2018transfer, shaw2019meta, Lian2020Towards, guobreaking}. \cite{chen2020catch} explores meta-learning to transfer network design knowledge from small tasks to larger tasks, surpassing many efficient solutions based on parameter sharing. \cite{cai2020tiny} proposes a highly memory-efficient and effective transfer solution that does not require back-propagation for adaptation. Central to these works' investigations are two key considerations: the transferability of an algorithm, namely how much information an algorithm can effectively reuse, and generalizability, which evaluates whether a solution can be applied to different settings (e.g., search spaces) and still performs well.

\begin{figure*}
\vspace{-3mm}
\begin{centering}
\includegraphics[width=2\columnwidth]{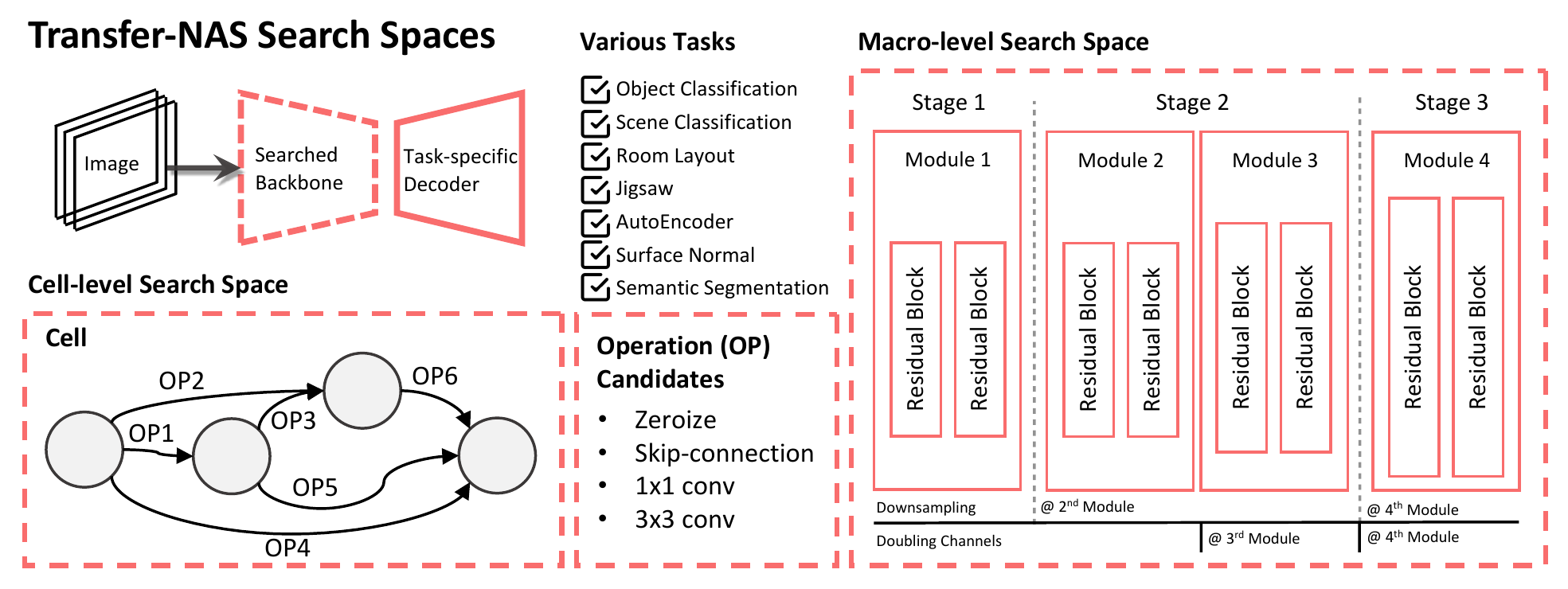}
\par\end{centering}
\caption{\label{fig:Our-Cell-level-and}Our cell-level and macro-level search space in TransNAS-Bench-101. We design a cell-level search space that treats each cell as a DAG and a macro-level search space that allows flexible macro skeleton network design.}
\vspace{-3mm}
\end{figure*}

Meanwhile, NAS is prohibitively costly. The computational intensity of single-task NAS can be discouraging, not to mention cross-task NAS experiments. To solve this computation limitation, NAS-Bench-101 \cite{ying2019bench} and NAS-Bench-201 \cite{dong2019bench} were proposed. These benchmarks have offered great values for the NAS community, but we believe the research scope of NAS can be further enlarged beyond classification problems and cell-based search spaces. To avoid confusion, throughout this paper we use \textbf{dataset} to refer to a set of original images, and \textbf{task} for pointing to certain vision domain. Hence, there can be multiple tasks for one dataset (e.g., COCO detection and segmentation).

The goal of finding universal solutions across tasks and search spaces, the comparability problem, and the computational barriers of transferable NAS research lead to our proposal of TransNAS-Bench-101, which studies networks over seven distinct vision tasks: object classification, scene classification, semantic segmentation, autoencoding, room layout, surface normal, and jigsaw. Two types of search spaces are provided: one is the macro skeleton search space based on residual blocks, where the network depth and block operations (e.g., where to raise the channel or downsample the resolution) are decided. Another one is the widely-studied cell-based search space, where each cell can be treated as a directed acyclic graph (DAG). The macro-level and cell-level search space contains 3,256 and 4,096 networks, respectively. The 7,352 backbones are evaluated on all seven tasks, and we provide detailed diagnostic information such as task performance, inference time, and FLOPs for all models. We also evaluated four transfer schemes compatible with both search spaces to serve as benchmarks for future research.

Our key contribution is a benchmark dataset with networks fully evaluated on seven tasks across two search spaces. Generating the benchmark takes 193,760 GPU hours, i.e., 22.12 years of computation on one NVIDIA V100 GPU, but it significantly lowers the cost of further research into cross-task neural architecture search. We also highlight problems and provide suggestions for future NAS research: (1) To extend NAS into different vision domains, it is important to look beyond cell-based search spaces, as we found that network macro structures can have bigger impact on performance on certain tasks. (2) The extent to which an algorithm surpasses random search is a crucial performance indicator. (3) Investigations of evolutionary-based transfer strategies, along with effective mechanisms to tweak transferred architectures and policies, are two promising directions for future research. With diversified settings in TransNAS-Bench-101, we hope to encourage the emergence of exceptional transferable NAS algorithms that generalize well under various settings. 

\section{The TransNAS-Bench-101 dataset}

\begin{figure*}[]
\vspace{-3mm}
\begin{centering}
\includegraphics[width=2\columnwidth]{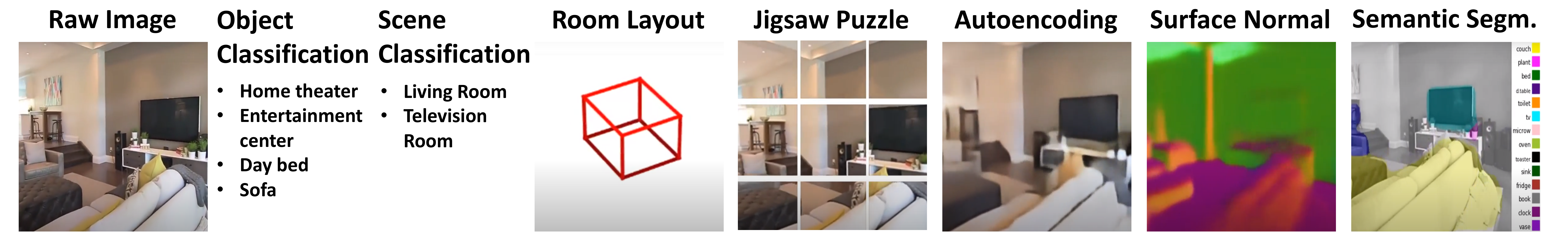}
\par
\caption{\label{fig:Tasks-considered-in}Vision tasks considered in our benchmarks.
We carefully select those 7 tasks above to ensure both diversity and similarity
across tasks from Taskonomy \cite{zamir2018taskonomy}.}
\end{centering}
\vspace{-2mm}
\end{figure*}

\begin{table*}[!b]
\small
\caption{\label{tab:Training-hyperparameters-and}Training hyperparameters
and details of each task considered in this benchmark. All the architectures
in the search space have been fully trained. We provide multiple metrics
for evaluation on the train/valid/test set. Each task requires a backbone-decoder network structure with task-specific decoder and loss function. GAP denotes global
average pooling. CE denotes the cross entropy loss. All the tasks except Autoencoding and Surface Normal use a cosine annealing with a linear warmup learning rate scheduler.}
\centering{}{\ \tabcolsep 0.000001in}%
\renewcommand\arraystretch{1.2}%
\begin{tabular}{ccccccccc}
\hline 
\textbf{\ Tasks} & \textbf{\ Decoder} & \textbf{\ LR} & \textbf{\ Optimizer} & \textbf{\ Output dim.} & \textbf{\ Loss} & \textbf{\ Eval. Metrics}\tabularnewline
\hline 
\textbf{\ Object Class.} & {\ GAP + Linear} & {\ 0.1}  & {\ SGD} & {\ 75} & {\ Softmax+CE} & {\ Loss, Acc}\tabularnewline
\textbf{\ Scene Class.} & {\ GAP + Linear} & {\ 0.1}  & {\ SGD} & {\ 47} & {\ Softmax+CE} & {\ Loss, Acc}\tabularnewline
\textbf{\ Room Layout} & {\ GAP + Linear} & {\ 0.1} & {\ SGD} & {\ 9} & {\ MSE loss} & {\ Loss}\tabularnewline
\textbf{\ Jigsaw} & {\ GAP + Linear} & {\ 0.1}  & {\ SGD} & {\ 1000} & {\ Softmax+CE} & {\ Loss, Acc}\tabularnewline
\textbf{\ Autoencoding} &{\ 14 Conv \& Deconv} & {\ 0.0005} & {\ Adam} & {\ 256x256} & {\ GAN loss+L1} & {\ Loss, SSIM}\tabularnewline
\textbf{\ Surface Normal} &{\ 14 Conv \& Deconv} & {\ 0.0001} & {\ Adam} & {\ 256x256} & {\ GAN loss+L1} & {\ Loss, L1, SSIM}\tabularnewline
\textbf{\ Sem. Segment.} & {\ 8 Conv \& Deconv} & {\ 0.1} & {\ SGD} & {\ 256x256} & {\ Softmax+CE} & {\ Loss, Acc, mIoU}\tabularnewline
\hline 
\end{tabular}{\scriptsize\par}
\vspace{-4mm}
\end{table*}

\subsection{Search Spaces and Architectures}

To plug in different networks for various tasks, our search space focuses on evolving the backbone, i.e., the mutual component of all the tasks considered. We provide two search spaces: a) A macro-level search space that designs the macro skeleton of a network, which was previously studied towards NAS in object detection and semantic segmentation; b) A cell-level search space following the widely studied cell-based search space, which applies to most weight-sharing NAS methods. 

\textbf{Macro-level Search Space. }Most NAS methods follow a fixed macro skeleton with a searched cell. However, the macro-level structure of the backbone can be crucial for the final performance. 
Early-stage feature maps in a backbone have larger sizes as they capture texture details, whereas feature maps at later stages are smaller and usually are more discriminative \cite{Li2018}. 
The allocation of computations over different stages is also vital for a backbone \cite{liang2019computation}.
Therefore, our search space contains networks with different depth (the total number of blocks), locations to down-sample feature maps, and locations to raise the channels. 
As is illustrated in Figure \ref{fig:Our-Cell-level-and}, we group two residual blocks \cite{he2016deep} into a module, and the networks are stacked with 4 to 6 modules. 
The module positions can be chosen to downsample the feature map 1 to 4 times, and each time the spatial size will shrink by a factor of 2. 
The network can double its channels 1 to 3 times at chosen locations. This search space thus consists of 3,256 unique architectures.

\begin{table*}[!t]
\caption{\label{tab:Related-work}Comparisons of TransNAS-Bench-101 with previous benchmarks. Although TransNAS-Bench-101 has a smaller search space, it contains more datasets, domains, and search space types.}
\vspace{-3mm}
\begin{center}
\begin{tabular}{ccccccc}
\hline
                   & \# Data- & \# Task & \# Search Space & Search Space\\ 
                   & sets     & Domains    & Size  & Type        \\ \hline
NAS-Bench-101      & 1        & 1          & 510M  & Cell       \\ 
NAS-Bench-201      & 3        & 1          & 15.6K & Cell      \\ 
TransNAS-Bench-101 & 1        & 7          & 7.3K   & Cell $\&$ Macro \\ \hline
\end{tabular}
\end{center}
\vspace{-5mm}
\end{table*}

\textbf{Cell-level Search Space.} We follow NAS-Bench-201\cite{dong2019bench}
to design our cell-level search space. It consists of 4,096 densely connected DAGs, which enables the evaluation of some weight-sharing NAS methods such as DARTS \cite{liu2018darts} and ProxylessNAS \cite{cai2018proxylessnas}.
As is shown in Figure \ref{fig:Our-Cell-level-and}, our cell-level search
space is obtained by assigning different operations (as edges) transforming
the feature map from the source node to the target node. The predefined
operation set has L=4 representative operations: zeroize, skip connection, 1x1 convolution, and 3x3 convolution. The \textit{convolution} in our setting represents an operation sequence of ReLU, convolution, and batch normalization. Each DAG consists of 4 nodes and 6 edges, including basic residual block-like cell designs. The macro-level skeleton is fixed, which contains five modules with doubling channel and down-sampling feature map operations at the 1st, 3rd, 5th modules.

Adding up the 3,256 and 4,096 networks from the macro-level search space and the cell-level search space, we have 7,352 unique architectures in total. All the architectures in both search spaces are carefully trained and evaluated across all the selected tasks.


\subsection{Dataset}

Unlike most NAS benchmarks that focus on classification tasks only, TransNAS-Bench-101 encourages the evaluation of algorithms across different tasks. Considered that Transferable NAS research is still in its infancy, we hope to provide as many common grounds for transfer as possible. This makes selecting proper datasets challenging since, ideally, the datasets should share some commonalities while covering a diversity of tasks.
Thanks to the great previous work Taskonomy \cite{zamir2018taskonomy}, which provides sufficient images with labels on different tasks (see Figure \ref{fig:Tasks-considered-in}), we can study algorithms without worrying that the datasets are by nature too distinct for transfer. The original dataset consists of 4.5M images of indoor scenes from about 600 buildings. To control the computational budget, we randomly select 24 buildings containing a total of 120K images from the original dataset and split the subset into 80K train / 20K val / 20K test set. 

\begin{table*}[b]
\small
\caption{\label{Table:dataset-comparison} We show the basic statistics of the two search spaces, including average performance metric scores, average model FLOPs, the average number of model parameters, and average training time. Note that here \textit{model} denotes the whole backbone-decoder model structure.}

\begin{centering}
\renewcommand\arraystretch{1.2}\tabcolsep 0.01in%
\begin{tabular}{l|c|ccccccc}
\hline 
\multicolumn{2}{c|}{{\ Tasks}} & {\ Cls. Object} & {\ Cls. Scene} & {\ Autoencoding} & {\ Surf. Normal} & {\ Sem. Segment.} & {\ Room Layout} & {\ Jigsaw} \tabularnewline
\hline 
\multicolumn{2}{c|}{{\ Metric}} & \textit{\ Acc.} & \textit{\ Acc.} & \textit{\ SSIM} & \textit{\ SSIM} & \textit{\ mIoU} & \textit{\ L2 loss} & \textit{\ Acc.} \tabularnewline
\hline 
  
 & { \ Performance} & { \ 39.71$\pm$5.92} & { \ 45.23$\pm$12.15} & { \ 0.46$\pm$0.10} & { \ 0.52$\pm$0.06} & { \ 19.95$\pm$5.52} & { \ -0.73$\pm$0.12} & { \ 76.57$\pm$28.34} \tabularnewline
 Cell & { \ FLOPs ($\times10^8$)} & { \ 2.44$\pm$1.44} & { \ 2.44$\pm$1.44} & { \ 4.90$\pm$1.44} & { \ 4.90$\pm$1.44} & { \ 5.19$\pm$1.44} & { \ 2.44$\pm$1.44} & { \ 1.38$\pm$0.81} \tabularnewline
 level & { \ Model Params ($\times10^6$)} & { \ 1.17$\pm$0.71} & { \ 1.17$\pm$0.71} & { \ 3.97$\pm$0.71} & { \ 3.97$\pm$0.71} & { \ 2.27$\pm$0.71} & { \ 1.17$\pm$0.71} & { \ 2.19$\pm$0.71} \tabularnewline
 & { \ Train Time (hr)} & { \ 2.08$\pm$0.23} & { \ 2.13$\pm$0.23} & { \ 5.03$\pm$0.54} & { \ 4.90$\pm$0.55} & { \ 5.98$\pm$1.69} & { \ 2.01$\pm$0.31} & { \ 1.28$\pm$0.19} \tabularnewline

\hline 
 & { \ Performance} & { \ 44.24$\pm$1.38} & { \ 52.92$\pm$2.08} & { \ 0.52$\pm$0.08} & { \ 0.57$\pm$0.02} & { \ 24.47$\pm$2.07} & { \ -0.65$\pm$0.03} & { \ 93.43$\pm$2.18} \tabularnewline
 Macro & { \ FLOPs ($\times10^8$)} & { \ 6.49$\pm$9.68} & { \ 6.49$\pm$9.68} & { \ 12.44$\pm$14.40} & { \ 12.44$\pm$14.40} & { \ 10.66$\pm$11.56} & { \ 6.49$\pm$9.68} & { \ 3.66$\pm$5.45} \tabularnewline
 level & { \ Model Params ($\times10^6$)} & { \ 1.18$\pm$0.91} & { \ 1.18$\pm$0.91} & { \ 3.83$\pm$1.02} & { \ 3.83$\pm$1.02} & { \ 2.13$\pm$1.02} & { \ 1.17$\pm$0.91} & { \ 1.91$\pm$1.14} \tabularnewline
 & { \ Train Time (hr)} & { \ 2.57$\pm$2.35} & { \ 2.56$\pm$2.25} & { \ 7.07$\pm$3.91} & { \ 7.10$\pm$4.02} & { \ 7.04$\pm$2.99} & { \ 2.59$\pm$2.25} & { \ 1.13$\pm$0.56} \tabularnewline

\hline 
\end{tabular}
\par\end{centering}
\vspace{-3mm}
\end{table*}

\subsection{Vision Tasks} 

We carefully selected seven tasks that lie on the intersection of (1) covering all major task categories in Taskonomy's task similarity tree \cite{zamir2018taskonomy}, and (2) align with the community's general research interests (e.g., classification tasks, or common and cheap pretrain tasks). As is shown in Figure \ref{fig:Tasks-considered-in}, the selected tasks include a) image classification tasks: Object Classification and Scene Classification; b) pixel-level prediction tasks: Surface Normal and Semantic Segmentation; c) self-supervised task: Jigsaw Puzzle and Autoencoding; d) point regression task: Room Layout.

\textbf{Object Classification.} Object classification is a 75-way classification problem that recognize objects. Labels provided by Taskonomy dataset \cite{zamir2018taskonomy} are activations generated by a ResNet-152 model \cite{he2016deep} pre-trained on ImageNet \cite{deng2009imagenet}. 

\textbf{Scene Classification. } Like object classification, scene classification is a 47-way classification problem that predicts the room type in the image. Its labels come from a ResNet-152 model pre-trained on MIT Places dataset \cite{zhou2017places}. Our selected dataset contains 47 classes out of the original 365 classes.

\textbf{Room Layout. } This task is to estimate and align a 3D bounding box defining the room layout. It requires a network to semantic information, such as "what constitutes a room," and includes scene geometry.

\textbf{Jigsaw.} We follow \cite{noroozi2016unsupervised} to design the self-supervised task Jigsaw. The input image is divided into nine patches and shuffled according to one of 1,000 preset permutations. The objective is to classify which permutation is used. This task's inclusion is inspired by a recent work \cite{liu2020labels} that explores neural architecture search without using labels.

\textbf{Autoencoding.} Autoencoding is a pixel-level prediction task that encodes images into low-dimensional latent representations then reconstructs the raw image. The training settings follow Conditional Adversarial Nets in Pix2Pix \cite{isola2017image}.

\textbf{Surface Normal. } Like autoencoding, surface normal is a pixel-wise prediction task that predicts surface normal statistics. The network structure and training procedure is the same as autoencoding. 

\textbf{Semantic Segmentation. } Semantic Segmentation conducts pixel-level prediction on the class of its components. The labels provided by the Taskonomy dataset are generated through a network pre-trained on the MSCOCO \cite{lin2014microsoft} dataset. Our selected subset contains 17 semantic classes.

\begin{figure*}[t]
    \centering
    \subfigure[Macro: FLOPs v. Rank]
    {
        \includegraphics[height=0.44\columnwidth]{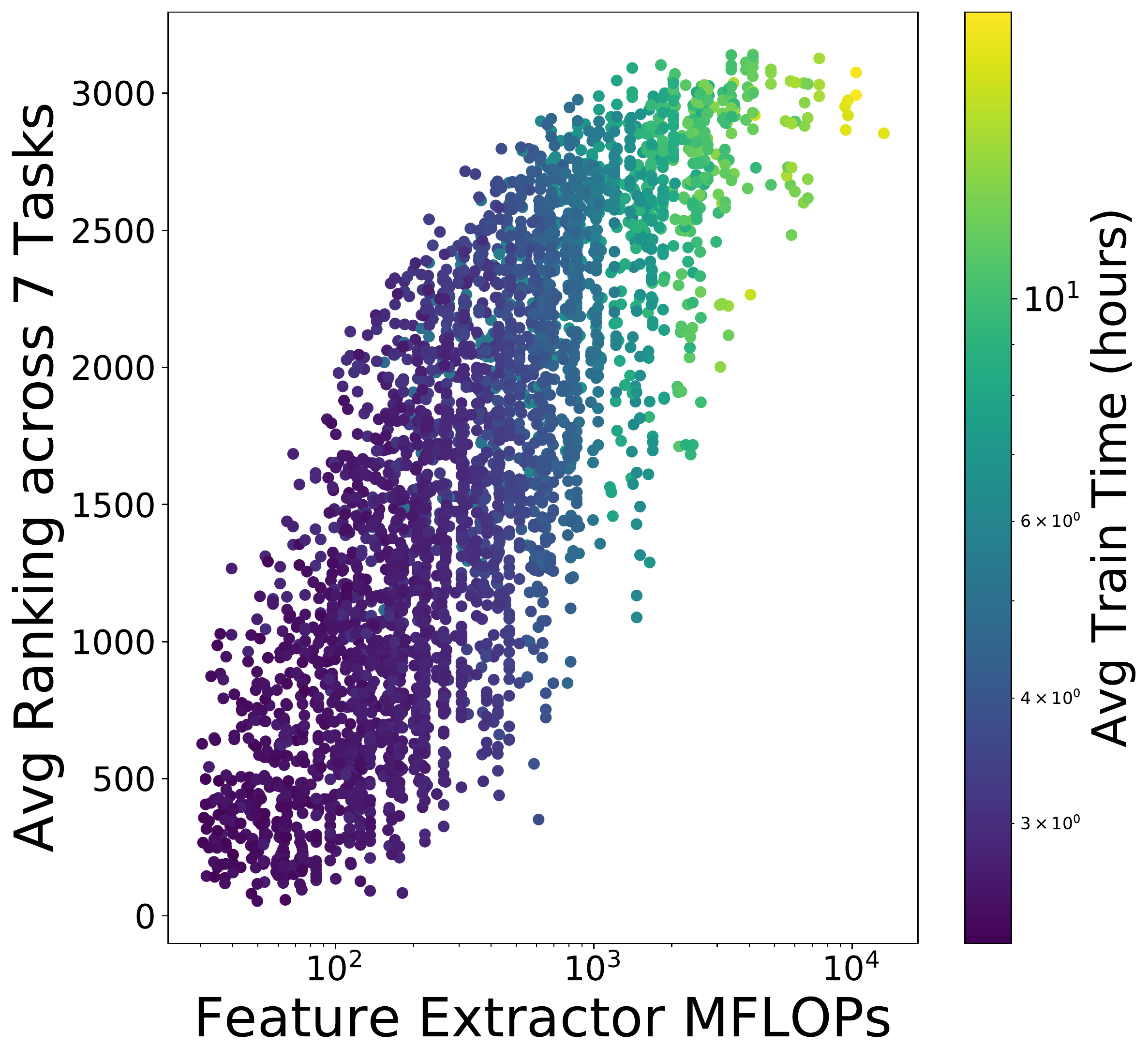}
        \label{fig:all-1}
    }
    \subfigure[Macro: Params v. Rank]
    {
        \includegraphics[height=0.44\columnwidth]{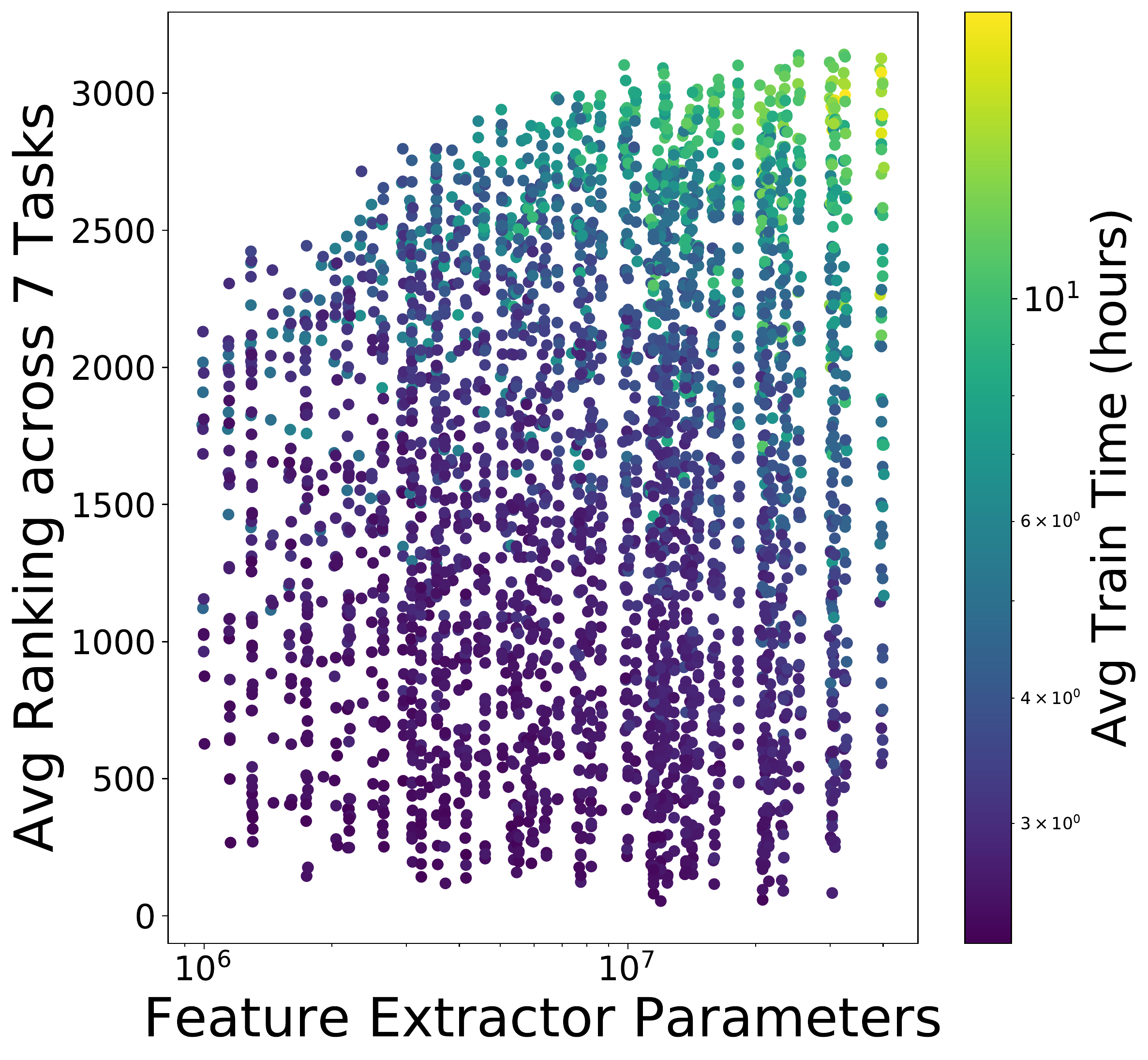}
        \label{fig:all-2}
    }
    \subfigure[Cell: FLOPs v. Rank]
    {
        \includegraphics[height=0.44\columnwidth]{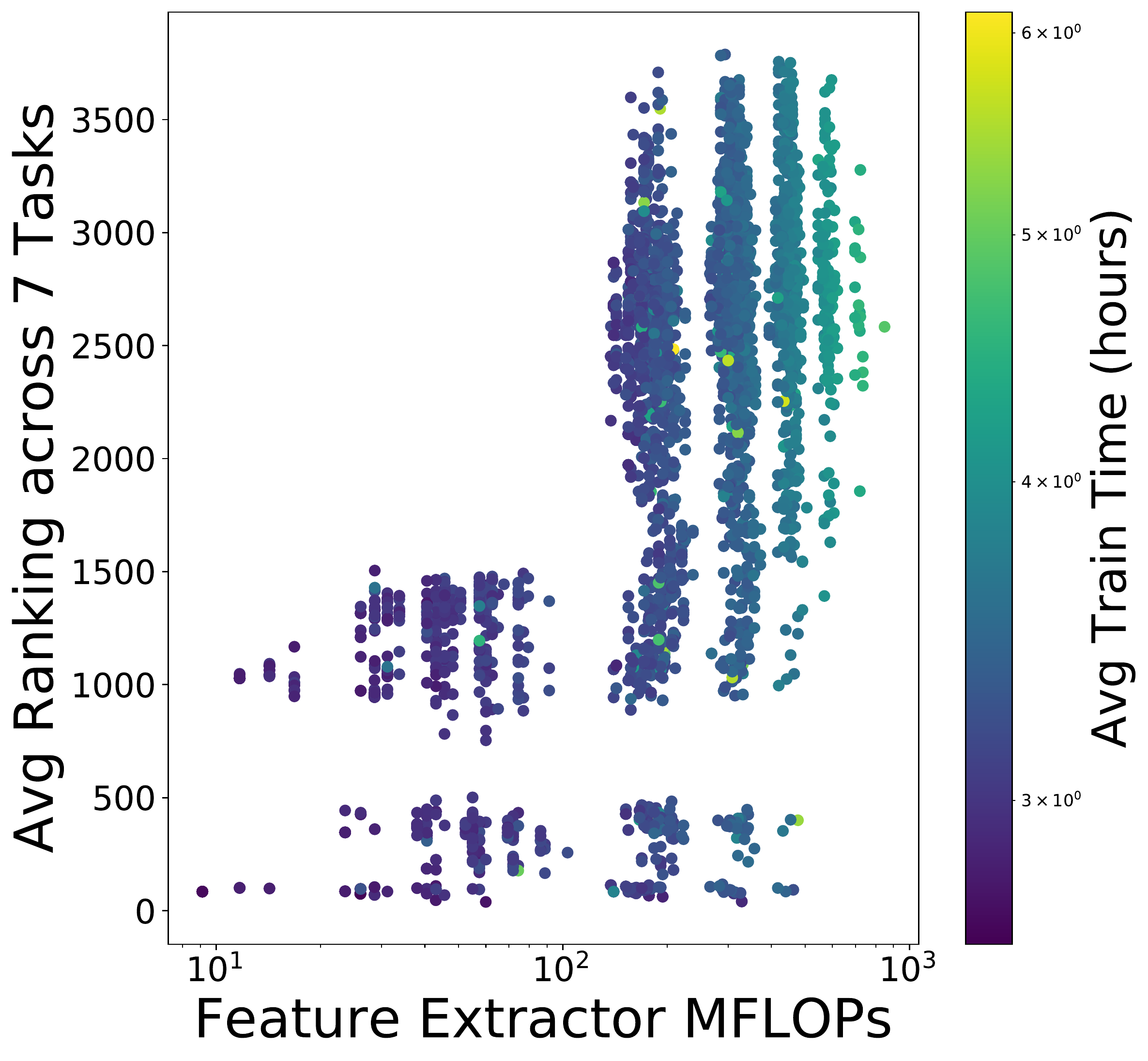}
        \label{fig:all-3}
    }
    \subfigure[Cell: Params v. Rank]
    {
        \includegraphics[height=0.44\columnwidth]{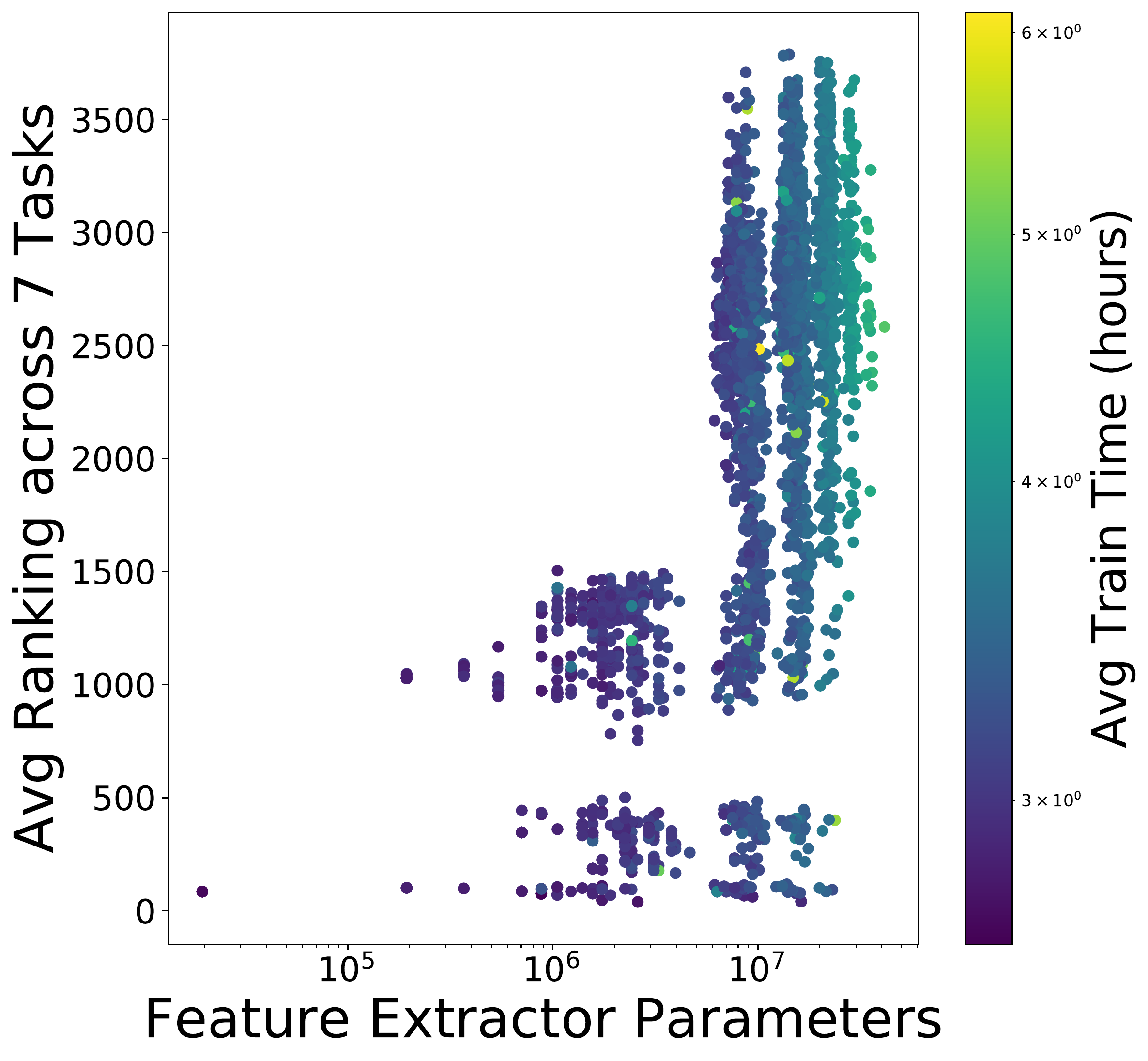}
        \label{fig:all-4}
    }
    \caption{The architecture average performance ranking, FLOPs, parameters, and training time on both search spaces. (a)-(b) display the overall landscape of the macro-level search space and (c)-(d) show the cell-level search space.}
    \label{fig:flops-and-time}
    \vspace{-3mm}
\end{figure*}

\begin{figure*}[!b]
    \vspace{-1mm}
    \centering
    \subfigure[Macro Correlation (all)]
    {
        \includegraphics[height=0.4\columnwidth]{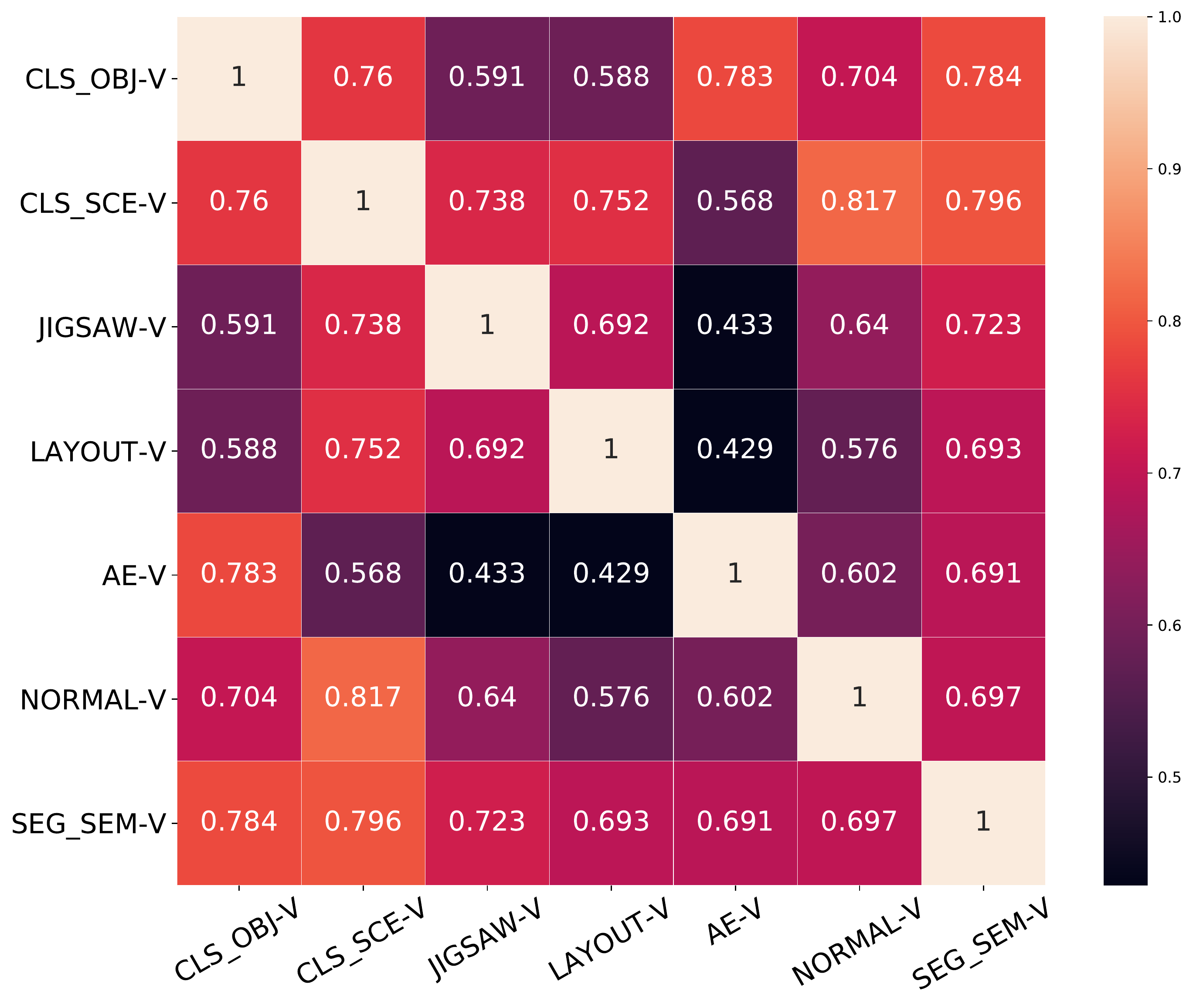}
        \label{fig:corr-1}
    }
    \subfigure[Macro Correlation (top 50\%)]
    {
        \includegraphics[height=0.4\columnwidth]{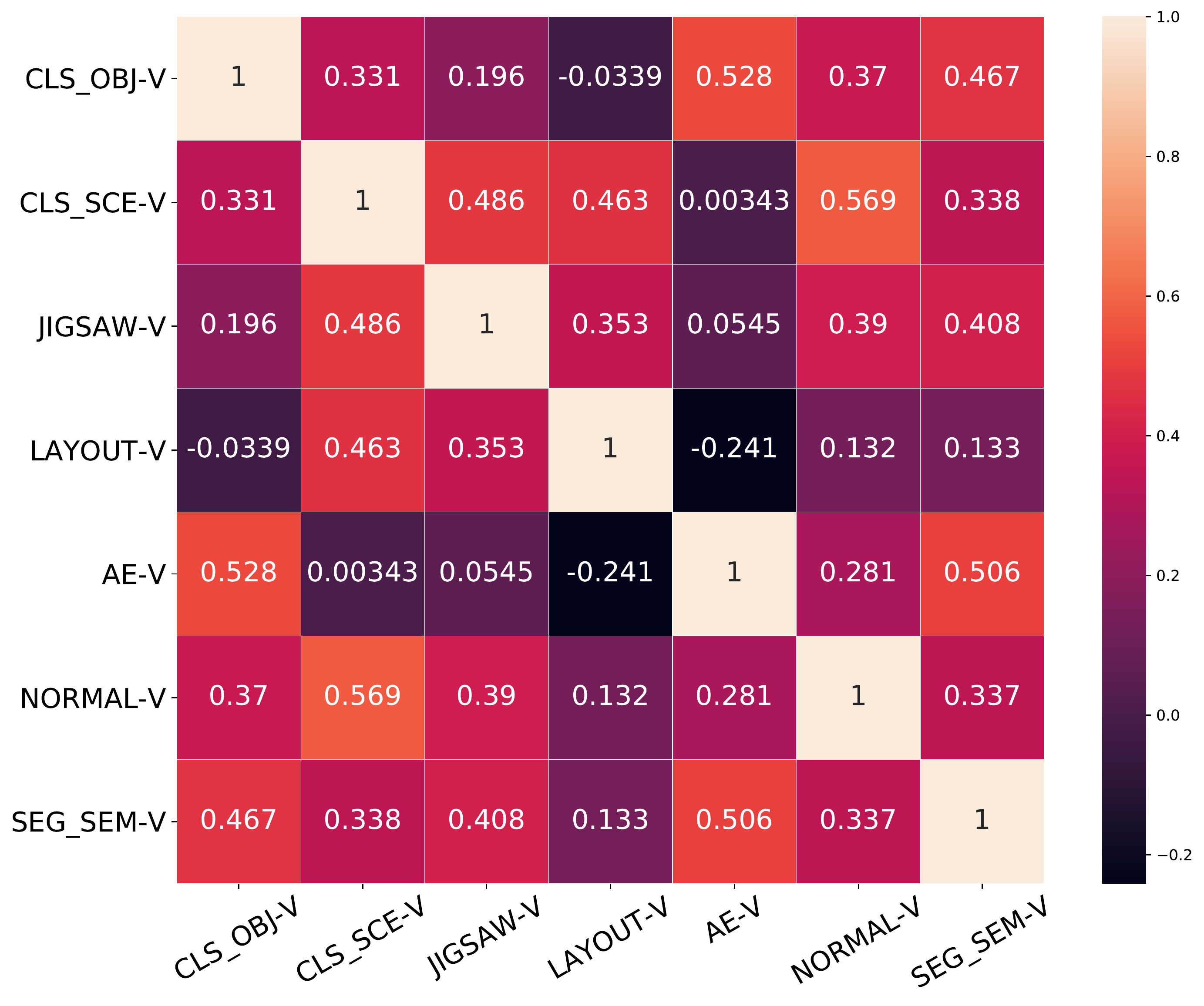}
        \label{fig:corr-2}
    }
    \subfigure[Cell Correlation (all)]
    {
        \includegraphics[height=0.4\columnwidth]{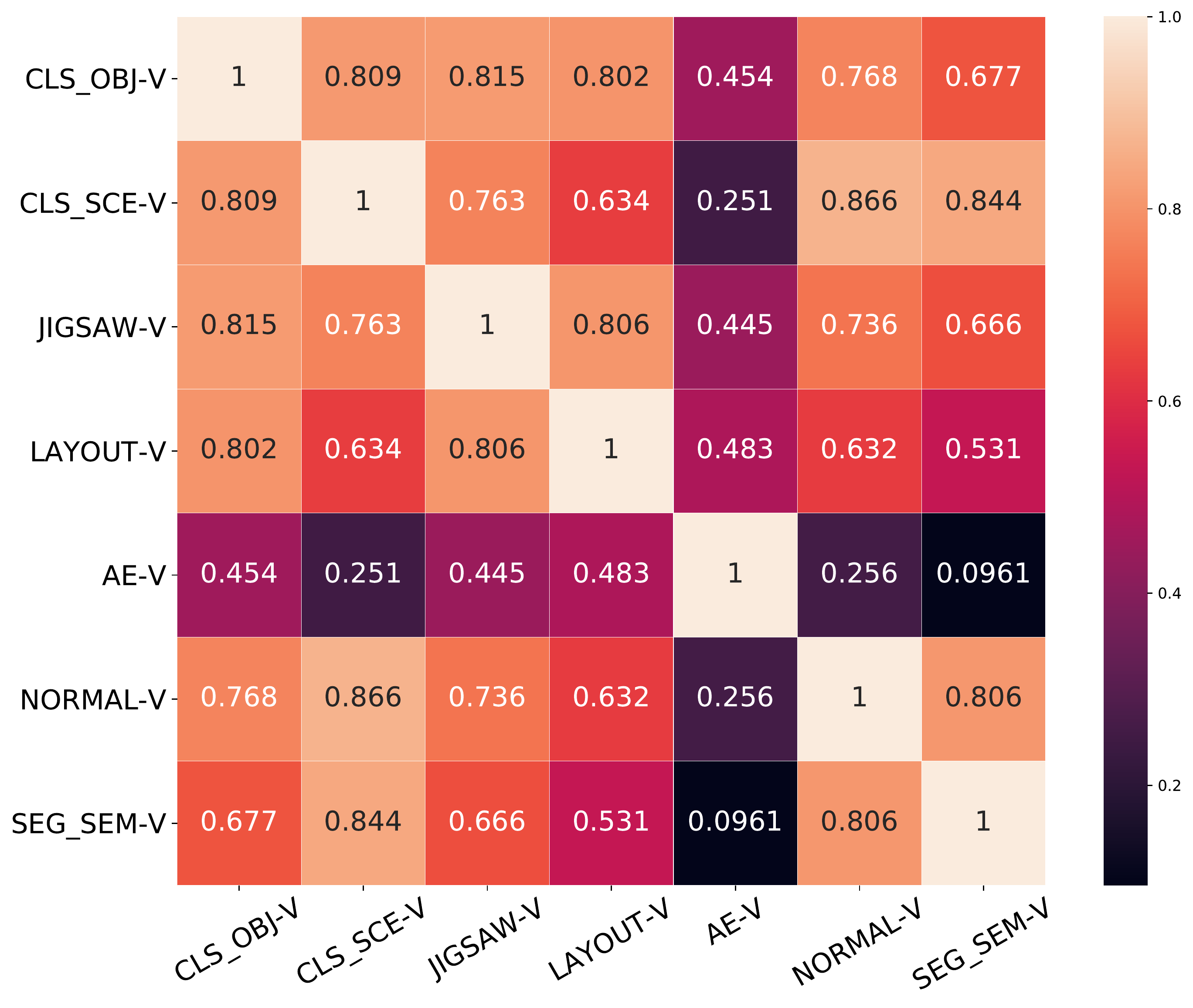}
        \label{fig:corr-3}
    }
    \subfigure[Cell Correlation (top 50\%)]
    {
        \includegraphics[height=0.4\columnwidth]{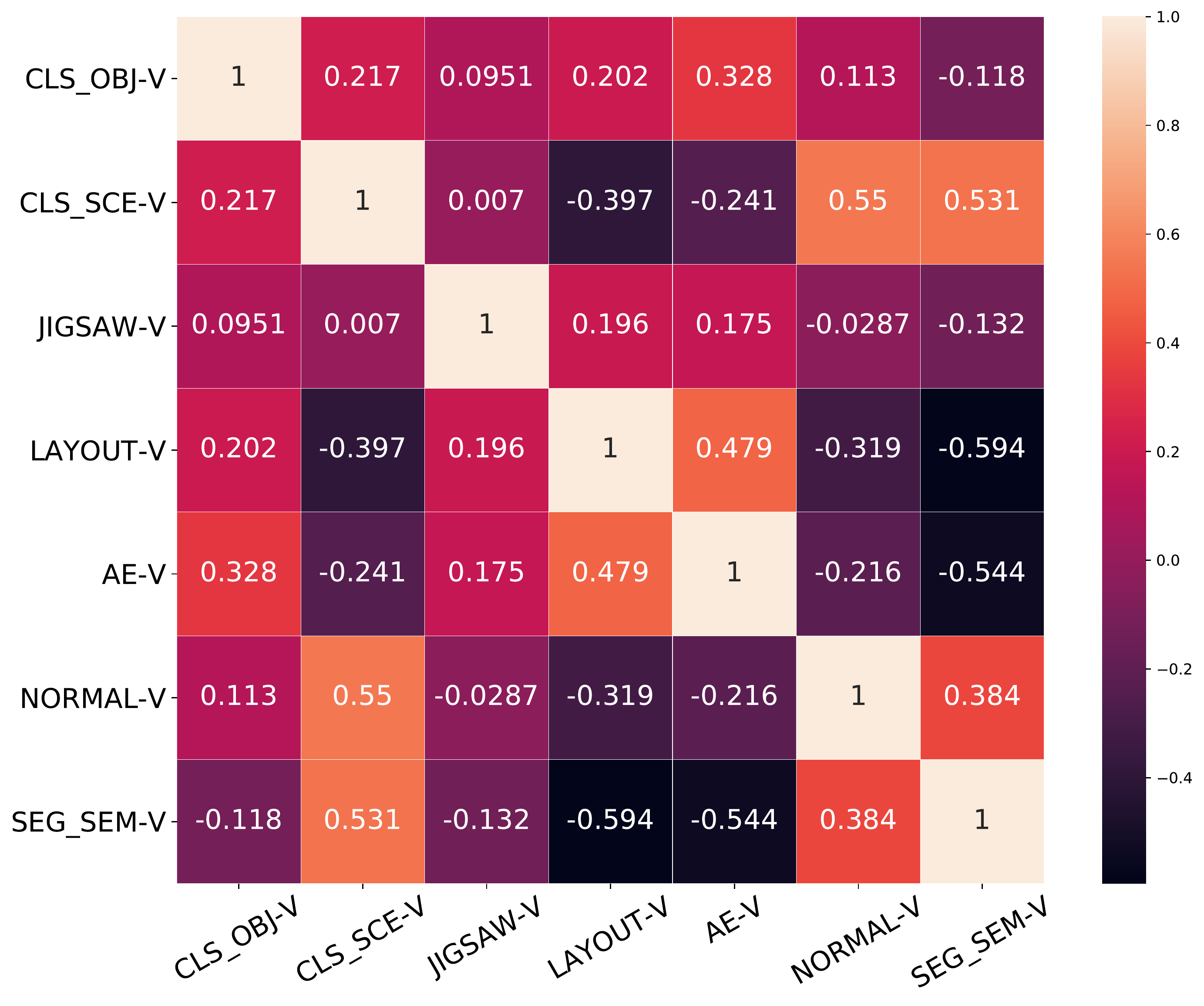}
        \label{fig:corr-4}
    }
    \caption{The Spearman rank correlations among tasks. (a), (c) show the correlation of all networks' validation performance scores across tasks on macro and cell search spaces, respectively. (b), (d) show the correlation of the top 50\% networks on macro and cell search spaces. Networks in the cell-level search space has higher correlations than the macro-level search space. The correlations shrinks quickly if we sample top 50\% of the networks.}
    \label{fig:correlation}
\end{figure*}

\subsection{Training Details\label{subsec:Training-Details}}

In TransNAS-Bench-101, the seven different tasks require different network structures and loss functions. To train the networks on a given task, we define a default backbone-decoder network structure first, then iterate through the search space and change its backbone architecture. For pixel-level prediction tasks and autoencoding, the decoders' input channels and resolutions will change flexibly but minimally according to different feature maps generated by the backbone. Since the original paper's implementation is based on an early version of Tensorflow, we reimplemented both the training and testing script with PyTorch for reproducibility. 

We mostly follow the Taskonomy paper to set up the hyper-parameters and training strategies shown in Table \ref{tab:Training-hyperparameters-and}.
For all the tasks, the batch size is 128, and the input resolution is resized to $256\times256$. We record multiple evaluation metrics in each epoch for all the architectures, as is listed in Table \ref{tab:Training-hyperparameters-and}.
Since we train every architecture in our search space for all the 7 tasks (i.e., $7352\times7\approx5\times10^{4}$ arch), the total computation cost is 193,760 GPU hours on V100 to generate the whole TransNAS-Bench-101.
Users can use our API to conveniently query each architecture's information across tasks without additional computation costs. In this way, researchers can significantly speed up their research and focus solely on improving the algorithms without tediously implementing and tuning different tasks.

\subsection{Network Information in TransNAS-Bench API}

We provide the train / validation / test performance information of each network at every epoch. The benchmark also contains each network's inference time, FLOPs, the total number of parameters, and time elapsed during each training epoch. Each model's inference time is measured on one Tesla V100 with one image of shape (3, 720, 1080). FLOPs are computed with one image of shape (3, 224, 224).

\section{Related Work}

To foster reproducibility and fair comparisons among algorithms, there are several existing NAS benchmarks. NAS-Bench-101 is the earliest work, which contains 423k unique architectures evaluated on the CIFAR-10 dataset. The networks are designed with a cell-based structure, where each cell is treated as a DAG. 

As an extension of NAS-Bench-101, NAS-Bench-201 was proposed to accommodate the growing needs. It provides training information about 15k networks, forming a complete search space. Similar to NAS-Bench-101, the networks are designed under a cell-based structure, but it could support many more algorithms with detailed diagnostic information. With training results over three datasets provided, it first enabled transfer learning across tasks. Ten benchmark algorithms are evaluated with extensive experiments in addition to network information. 

TransNAS-Bench-101's commonalities with previous benchmarks mainly lie in: (1) It offers detailed network training information with all the networks in an entire search space. (2) It also adopts the cell-based search space, treating each cell as a DAG. However, TransNAS-Bench-101 evaluates its networks across a much more diversified set of tasks. It is also the first benchmark that provides a thorough analysis of the macro skeleton search space.

\section{Analysis of TransNAS-Bench-101}

\subsection{Overview of architectures} 

The architectures' performance ranking, FLOPs, parameters, and training time are presented in Figure \ref{fig:flops-and-time}. We obtain rankings of each architecture's validation performance on all 7 tasks first, then plot the average ranking. A higher number means a better performance ranking.

The pattern shows that a network with more FLOPs and longer training time tends to perform better on the given tasks within a reasonable range, but it does not include some of the search space's largest networks. Figure \ref{fig:flops-and-time} also reveals some distinctive characteristics of both search spaces. The macro-level search space has its networks more evenly spread out in terms of FLOPs, whereas networks in the cell-level search space are more concentrated at certain numbers. The macro-level search space's network FLOPs vary across a wider range from $10^8$ to $10^{11}$, while the cell-level search space's architectures range across $10^8$ to $10^{10}$, which is a magnitude smaller. The FLOPs and parameter patterns are similar in the cell-level search space, but the pattern significantly changed when we investigate the macro-level search space. It can take up to 12 hours to train a network in the macro-level search space, which is twice the GPU hour needed to train the most computationally demanding network on the cell-level search space.

\begin{figure}[t]
    \centering
    \subfigure[]
    {
        \includegraphics[height=0.35\columnwidth]{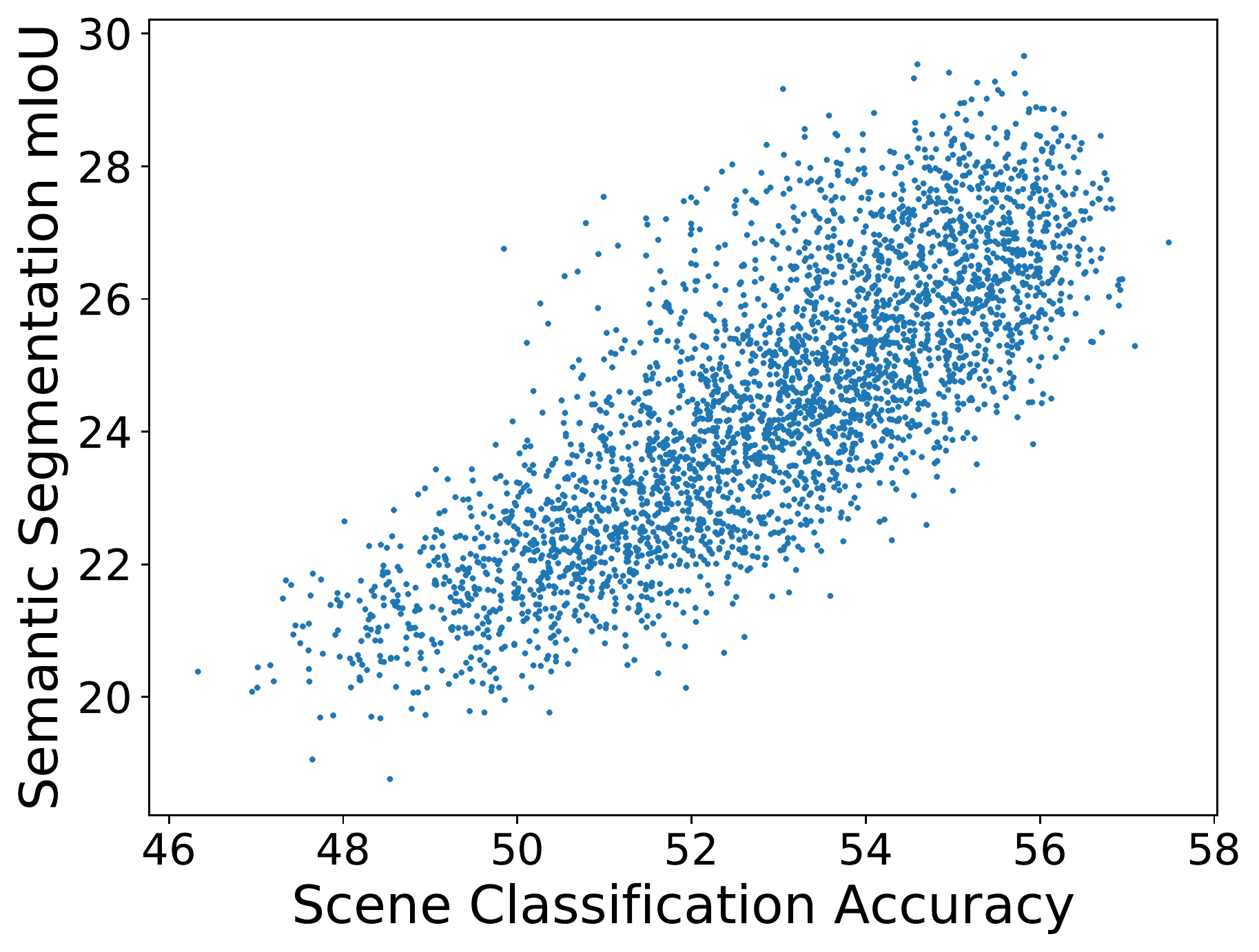}
    }
    \subfigure[]
    {
        \includegraphics[height=0.35\columnwidth]{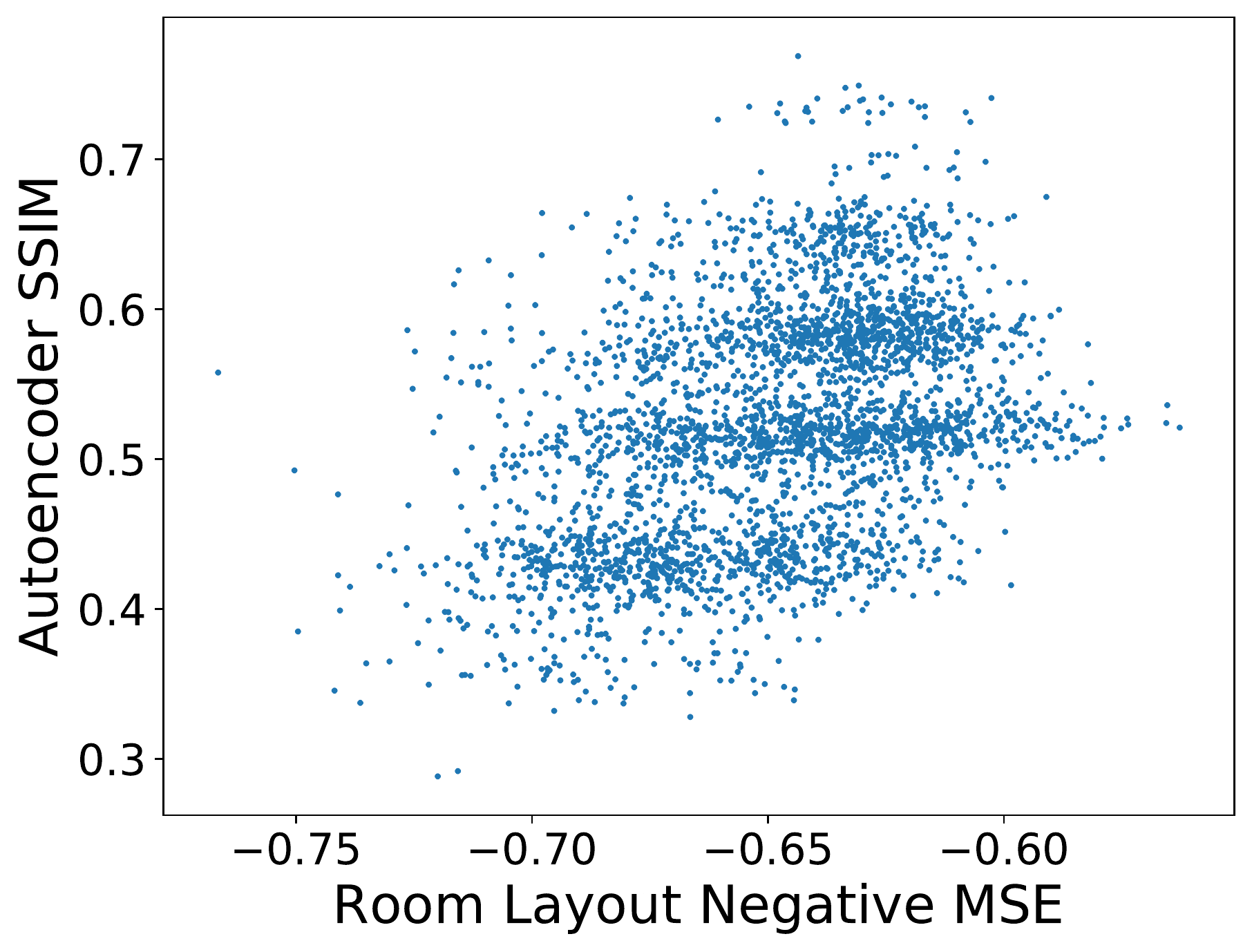}
    }
    \caption{Tasks with the highest and lowest correlations on the macro-level search space.}
    \label{fig:scene-seg-ae-rl}
    \vspace{-3mm}
\end{figure}

\subsection{Correlations among tasks} 

We analyze cross-task correlations by calculating the Spearman's Rank Correlation coefficient among the tasks and present the results in Figure \ref{fig:correlation}. Although object classification and scene classification are both classification tasks, scene classification has higher correlations with surface normal and semantic segmentation tasks on both search spaces than object classification. This phenomenon shows that tasks within the same domain, even though they are based on essentially the same images, might not necessarily be closer in terms of architectural performance. We visualize the network performance of tasks with the highest (0.817) and lowest (0.429) correlations on the macro-level search space in Figure \ref{fig:scene-seg-ae-rl}.

The Autoencoding task has very distinctive behaviors under the two search spaces. Similar to semantic segmentation and surface normal, it is an image translation task that outputs 256x256 images. With networks in the macro-level search space, the autoencoding task has moderate correlations with semantic segmentation (0.691) and surface normal (0.602). However, under the cell-level search space, it has almost no correlations with semantic segmentation (0.0961), and very weak correlation with surface normal (0.256). This considerable discrepancy shows that the selection of search space can significantly impact specific tasks. Some search spaces can dramatically lower the difficulty of NAS transfer, and some might have inherent disadvantages. It again highlights the importance of validating an algorithm's performance on multiple search spaces to obtain unbiased evaluations. 

The correlations among tasks shrink quickly if we plot the graphs with only the top 50\% networks' performance information. Some tasks still have relatively strong correlations, but others rapidly drop to below zero. This shows that the direct transfer strategy of architectures might not always yield good results, and robustly transferable algorithms should be wary of it to avoid negative transfer. 


\begin{figure*}[!t]
    \centering
    \subfigure[Macro Search Space on Room Layout]
    {
        \includegraphics[height=0.46\columnwidth]{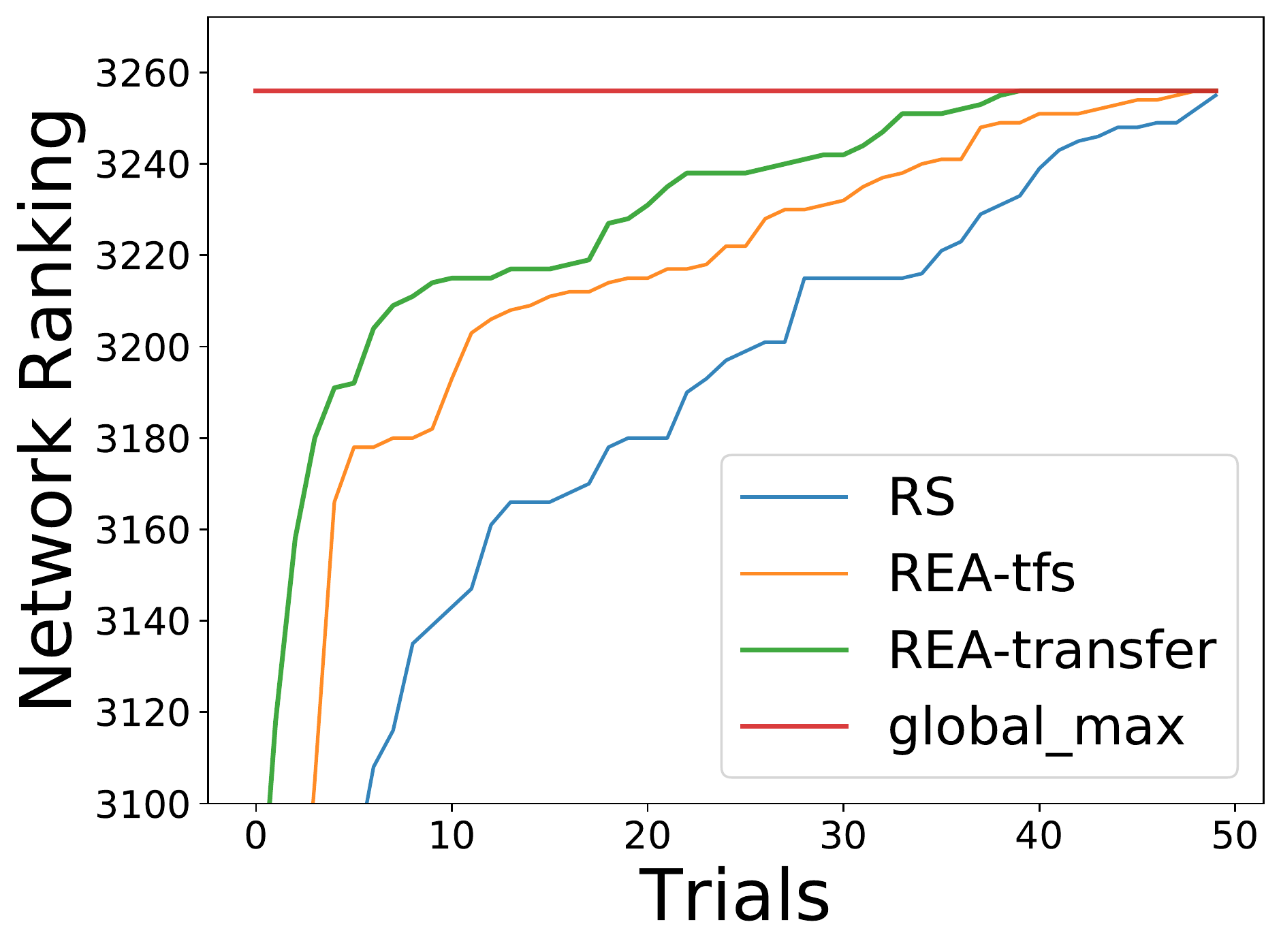}
        \label{fig:rea-1}
    }
    \subfigure[REA on Macro Search Space (Averaged)]
    {
        \includegraphics[height=0.46\columnwidth]{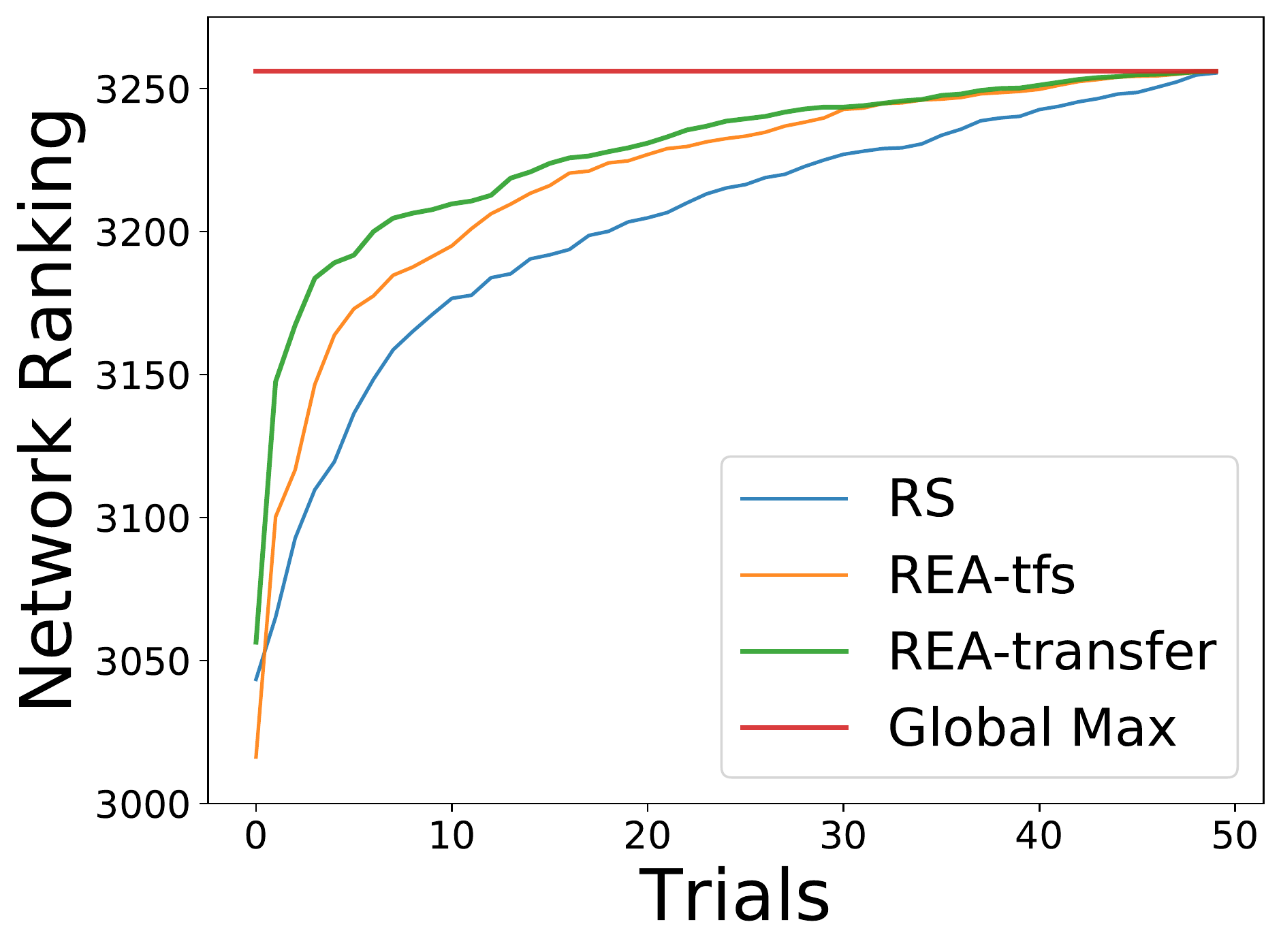}
        \label{fig:rea-2}
    }
    \subfigure[REA on Cell Search Space (Averaged)]
    {
        \includegraphics[height=0.46\columnwidth]{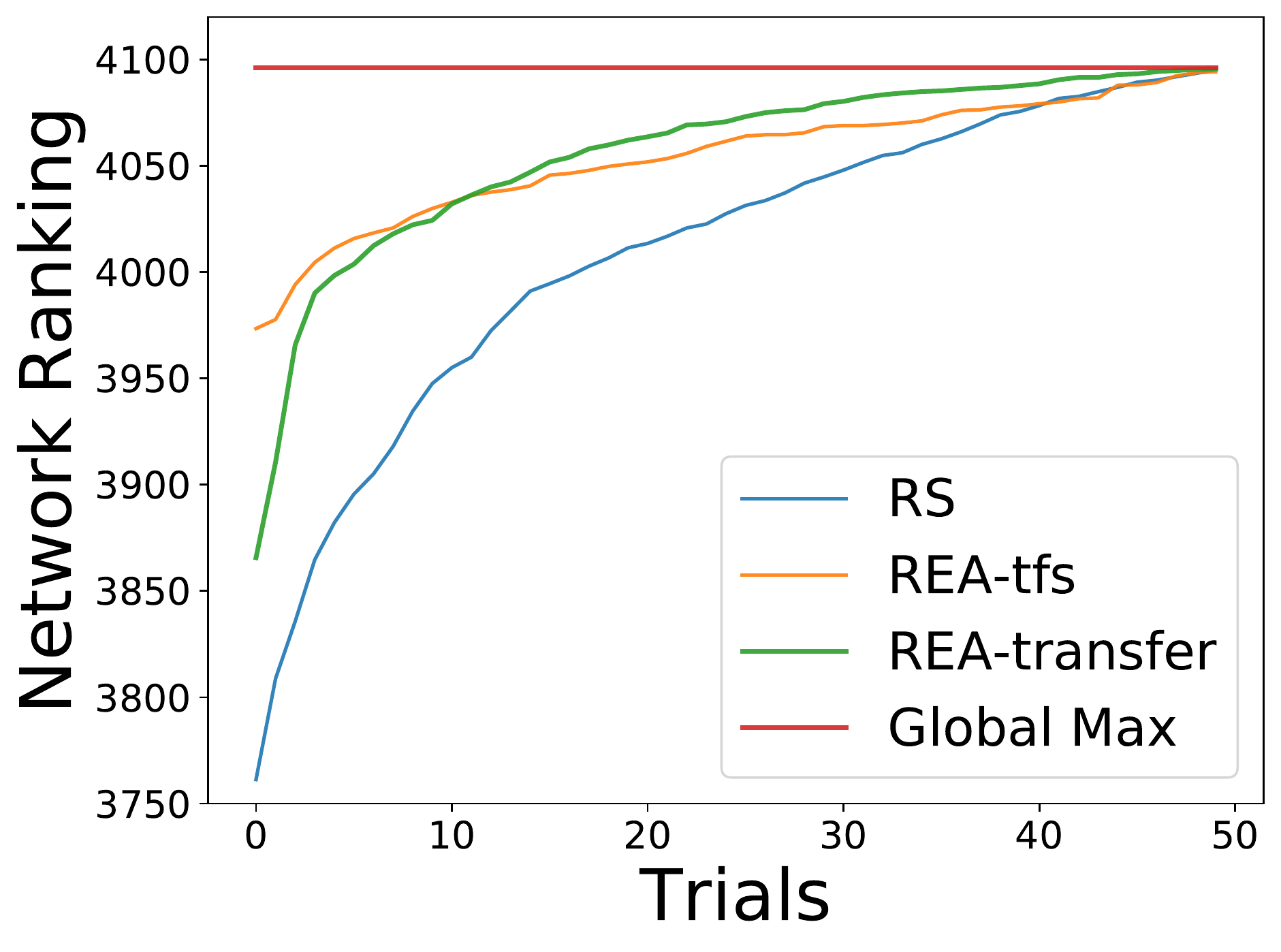}
        \label{fig:rea-3}
    }
    \caption{Comparisons of the transfer and train-from-scratch (tfs) results of REA. REA-transfer has slight but stable improvements across all tasks. We ran each algorithm in \ref{fig:rea-1} for 50 trials and each trial to search 50 networks. We pick the model with the highest validation score in history as the search result of a trial. \ref{fig:rea-1} shows sorted search results of on Room Layout. Each curve in \ref{fig:rea-2} and \ref{fig:rea-3} averages the search results across all tasks.}
    \label{fig:rea}
    \vspace{-3mm}
\end{figure*}

\begin{figure*}[t]
    \centering
    \subfigure[Macro Search Space on Object Class.]
    {
        \includegraphics[height=0.46\columnwidth]{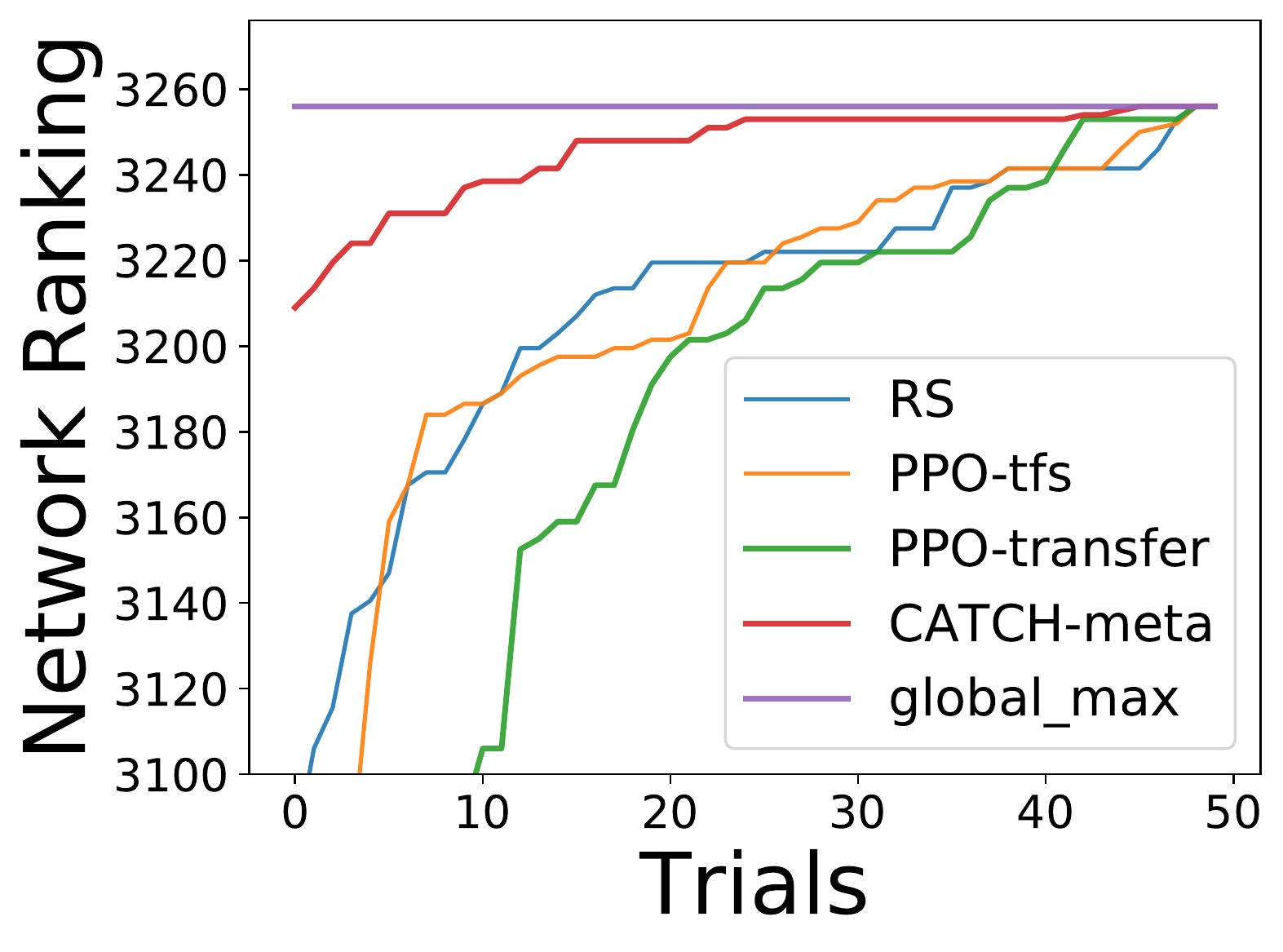}
        \label{fig:rl-1}
    }
    \subfigure[RL on Macro Search Space (Averaged)]
    {
        \includegraphics[height=0.46\columnwidth]{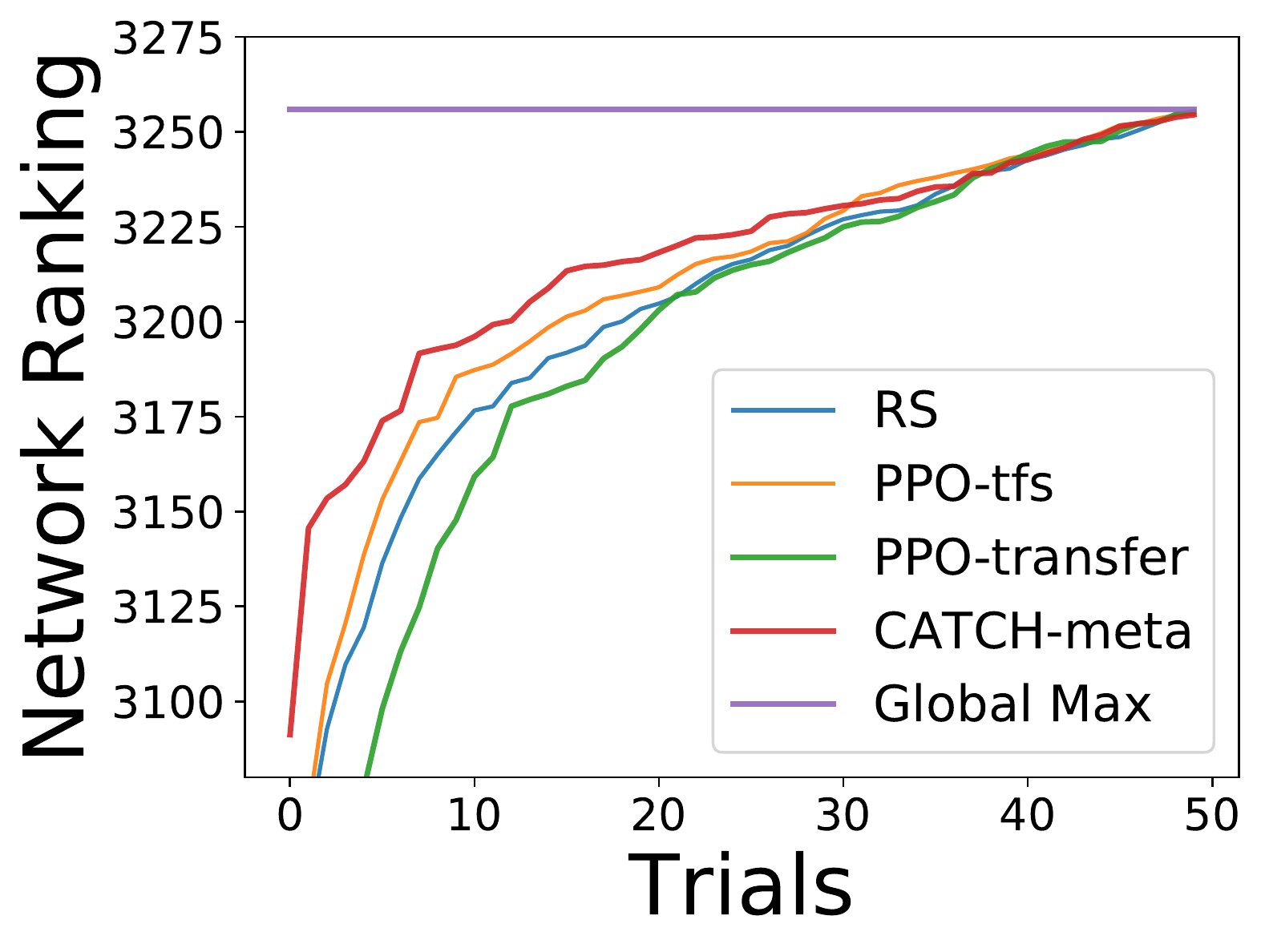}
        \label{fig:rl-2}
    }
    \subfigure[RL on Cell Search Space (Averaged)]
    {
        \includegraphics[height=0.46\columnwidth]{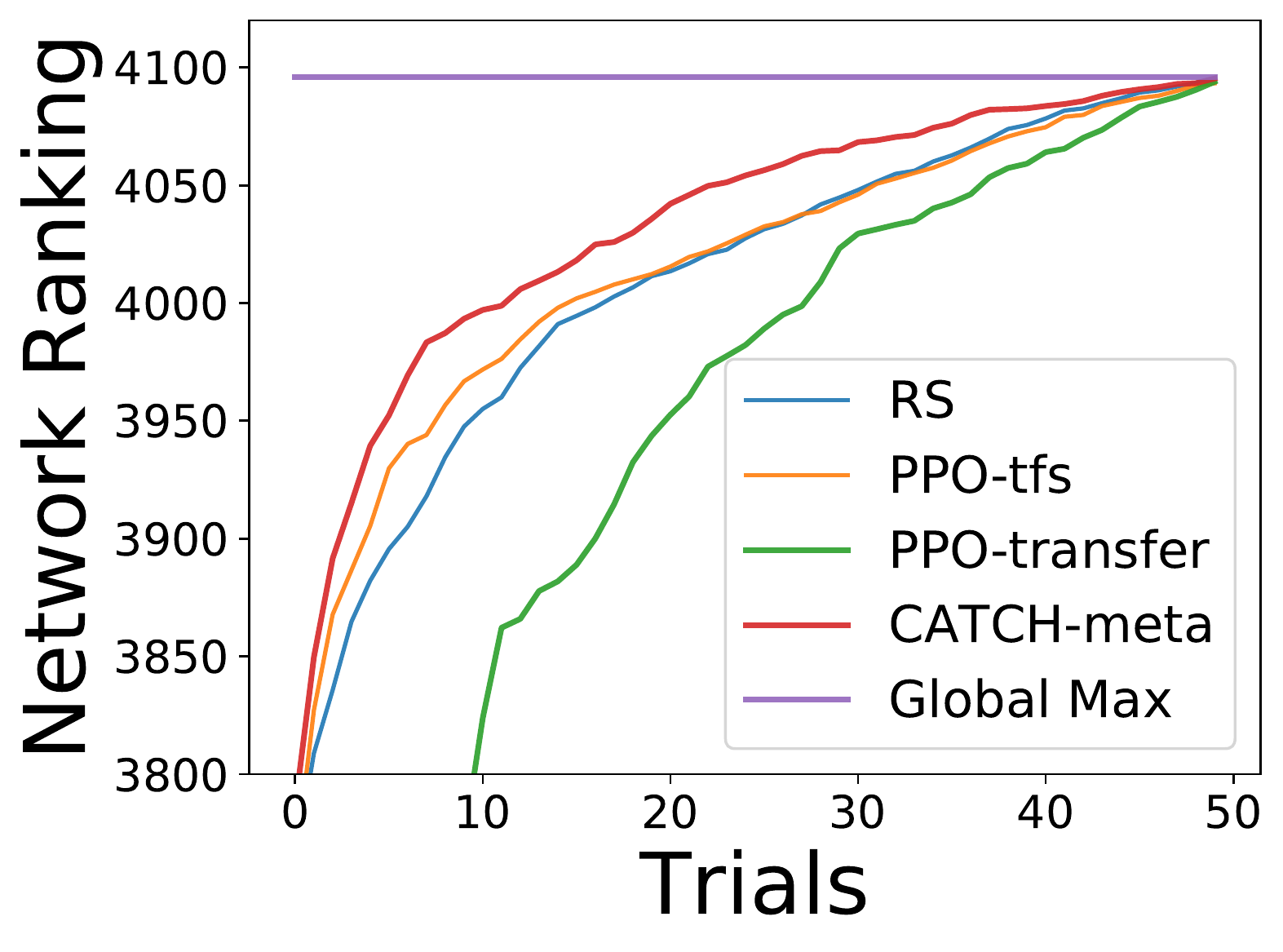}
        \label{fig:rl-3}
    }
    \caption{Comparison of PPO and CATCH. CATCH largely improves PPO-transfer’s performance, and it works exceedingly well on object classification. Similar to REA, We ran each algorithm in \ref{fig:rl-1} for 50 trials and each trial to search 50 networks. \ref{fig:rl-1} shows sorted search results on object classification. Each curve in \ref{fig:rl-2} and \ref{fig:rl-3} averages search results across all tasks.}
    \label{fig:rl}
    \vspace{-3mm}
\end{figure*}

\section{Benchmark Algorithms}

Because the field of transferable NAS is new and nascent, there are only limited number of transferable architecture search algorithms. We implemented four baseline algorithms using TransNAS-Bench-101 and provide  implementation details and results in this section.

We evaluated 5 transfer schemes of different types: (1) Random Search (RS) \cite{bergstra2012random}; (2) Direct transfer of top architectures (DT); (3) Direct policy transfer of reinforcement learning-based algorithm, e.g., Proximal Policy Optimization (PPO) \cite{schulman2017proximal}; (4) Meta-learning based algorithm, e.g., CATCH  \cite{chen2020catch}; (5) Evolutionary algorithms with transferred population initialization, e.g., Regularized Evolutionary Algorithm (REA) \cite{real2019regularized}. Each selected algorithm represents a distinctive type of transfer scheme, and they are compatible with both search spaces. The algorithms are tested using the MindSpore tools \cite{Mindspore}.

\textbf{Evaluation of Random Search and Direct Transfer}. Many NAS algorithms have attempted to use architectures as the central component for transfer \cite{zoph2018learning, pasunuru2019continual}, which inspired us to look into the efficiency of direct transfer of top architectures found by RS. For both search spaces, we randomly search 50 architectures in each trial. The architecture with the highest validation accuracy will be selected as the final searched result of this particular trial. 
In Table \ref{Table:comparison} we include the results for direct transfer (DT). For each task, we take the best architecture from the remaining 6 tasks, and average their performance on the selected task. The results are significantly worse than all the search algorithms. It shows that even if some NAS algorithm can be guaranteed to find the optimal architecture on a source task, its direct transfer performance on the target task can still suffer.

\begin{table*}[t]
\caption{\label{Table:comparison}Performance comparison of different transferable NAS methods. Room layout's L2 loss is multiplied by a factor of 100 for better readability. The transferred versions of REA and PPO are pretrained on the least time-consuming task, Jigsaw. The rightmost column reports the percentile of searched networks (Perc.) in the benchmark, averaged across all target tasks (i.e. all tasks except Jigsaw).}

\begin{centering}
\small
\renewcommand\arraystretch{1.2}\tabcolsep 0.03in%
\begin{tabular}{l|c|ccccccc|c}
\hline 
\multicolumn{2}{c|}{{Tasks}} & {Cls. Object} & {Cls. Scene} & {Autoencoding} & {Surf. Normal} & {Sem. Segment.} & {Room Layout} & {Jigsaw} & {Perc.}\tabularnewline
\hline 
\multicolumn{2}{c|}{{Metric}} & \textit{Acc.} & \textit{Acc.} & \textit{SSIM} & \textit{SSIM} & \textit{mIoU} & \textit{L2 loss} & \textit{Acc.} & \tabularnewline
\hline 
 & {RS} & {45.16$\pm$0.4} & {54.41$\pm$0.3} & {55.94$\pm$0.8} & {56.85$\pm$0.6} & {25.21$\pm$0.4} & {61.48$\pm$0.8} & {94.47$\pm$0.3} & {97.91}\tabularnewline
 & {REA-tfs} & {45.39$\pm$0.2} & {54.62$\pm$0.2} & {56.96$\pm$0.1} & {57.22$\pm$0.3} & {25.52$\pm$0.3} & {61.75$\pm$0.8} & {94.62$\pm$0.3} & {99.06}\tabularnewline
{Cell} & {REA-transfer} & {45.51$\pm$0.3} & {54.61$\pm$0.2} & {56.52$\pm$0.6} & {57.20$\pm$0.7} & {25.46$\pm$0.4} & {61.04$\pm$1.0} & {-} & {99.02}\tabularnewline
{level} & {PPO-tfs} & {45.19$\pm$0.3} & {54.37$\pm$0.2} & {55.83$\pm$0.7} & {56.90$\pm$0.6} & {25.24$\pm$0.3} & {61.38$\pm$0.7} & {94.46$\pm$0.3} & {98.07}\tabularnewline
 & {PPO-transfer} & {44.81$\pm$0.6} & {54.15$\pm$0.5} & {55.7.0$\pm$1.5} & {56.60$\pm$0.7} & {24.89$\pm$0.5} & {62.01$\pm$1.0} & {-} & {94.94}\tabularnewline
 & {CATCH} & {45.27$\pm$0.5} & {54.38$\pm$0.2} & {56.13$\pm$0.7} & {56.99$\pm$0.6} & {25.38$\pm$0.4} & {60.70$\pm$0.7} & {-} & {98.45}\tabularnewline
 & {DT} & {\ 42.03$\pm$4.96 } & {\ 49.80$\pm$8.59 } & {\ 51.20$\pm$3.32 } & {\ 55.03$\pm$2.68 } & {\ 22.45$\pm$3.24 } & {\ 66.98$\pm$2.25 } & {\ 88.95$\pm$9.13 } & {\ 77.17 }\tabularnewline
\cline{2-10} \cline{3-10} \cline{4-10} \cline{5-10} \cline{6-10} \cline{7-10} \cline{8-10} \cline{9-10} \cline{10-10}
 & {Global Best} & {46.32} & {54.94} & {57.72} & {59.62} & {26.27} & {59.38} & {95.37} & {100}\tabularnewline
\hline 
 & {RS} & {46.85$\pm$0.3} & {56.5$\pm$0.4} & {70.06$\pm$3.1} & {60.70$\pm$0.9} & {28.37$\pm$0.5} & {59.35$\pm$1.0} & {96.78$\pm$0.2} & {98.18} \tabularnewline
 & {REA-tfs} & {47.09$\pm$0.4} & {56.57$\pm$0.4} & {69.98$\pm$3.6} & {60.88$\pm$1.0} & {28.87$\pm$0.4} & {58.73$\pm$1.1} & {96.88$\pm$0.2} & {98.74} \tabularnewline
{Macro} & {REA-transfer} & {46.98$\pm$0.4} & {56.60$\pm$0.3} & {73.41$\pm$2.9} & {61.02$\pm$0.8} & {28.90$\pm$0.5} & {58.18$\pm$1.3} & {-} & {99.11} \tabularnewline
{level} & {PPO-tfs} & {46.84$\pm$0.4} & {56.48$\pm$0.3} & {70.92$\pm$3.2} & {60.82$\pm$0.8} & {28.31$\pm$0.5} & {58.84$\pm$1.1} & {96.76$\pm$0.2} & {98.33} \tabularnewline
 & {PPO-transfer} & {46.76$\pm$0.5} & {56.47$\pm$0.4} & {70.54$\pm$2.9} & {60.80$\pm$1.3} & {28.31$\pm$0.6} & {59.17$\pm$0.8} & {-} & {97.79} \tabularnewline
 & {CATCH} & {47.29$\pm$0.3} & {56.49$\pm$0.3} & {70.36$\pm$3.0} & {60.85$\pm$0.7} & {28.71$\pm$0.4} & {59.37$\pm$0.6} & {-} & {98.78} \tabularnewline
  & {DT} & {\ 45.48$\pm$1.02 } & {\ 54.96$\pm$1.80 } & {\ 59.35$\pm$8.99 } & {\ 58.60$\pm$1.56 } & {\ 26.21$\pm$1.91 } & {\ 62.07$\pm$1.43 } & {\ 95.37$\pm$1.55 } & {\ 83.59 } \tabularnewline 
\cline{2-10} \cline{3-10} \cline{4-10} \cline{5-10} \cline{6-10} \cline{7-10} \cline{8-10} \cline{9-10} \cline{10-10} 
 & {Global Best} & {47.96} & {57.48} & {76.88} & {64.35} & {29.66} & {56.28} & {97.02} & {100}\tabularnewline 
\hline 
\end{tabular}
\par\end{centering}
\vspace{-3mm}
\end{table*}

\textbf{Evaluation of policy transfer. }
For both search spaces, the NAS problem can be formulated as a sequential decision problem, and the reinforcement learning algorithm PPO aims to select each attribute choice to form a network. We use a multi-layer perceptron (MLP) as the policy network for PPO, and pre-train the policy for 50 epochs on a less time-costly source task (Jigsaw in our case). The pre-trained policy is then adapted to the target task to search for another 50 epochs. 

CATCH uses PPO as its controller to sample networks. It also uses an MLP network evaluator to predict the network performance and uses a context encoder to learn a task-specific embedding to guide the search. CATCH incorporated meta reinforcement learning by first meta-training the policy on low-cost tasks and then adapt the meta-trained policy to a target task. We use jigsaw, object classification, and scene classification tasks as its meta-training tasks. CATCH adopts the same budget for pre-train and transfer, and we repeat each trial 50 times. We plot the search results on each target task and then take the average of the curves across all tasks. The result is shown in Figure \ref{fig:rl}. 

The direct transfer of PPO policies shows worse results than its non-transfer version, a phenomenon commonly referred to as negative transfer. We conjecture that this is due to PPO's overfitting to the source task during the pre-train phase. CATCH mediates it with two added components: An encoder that provides task information that guides its policy and an evaluator that filters inferior candidates. From our experiments, these added components do improve the transfer results under certain circumstances. As Table \ref{Table:comparison} indicates, it shows exceedingly good performance under certain settings such as object classification on the macro-level search space, but it also struggles on some other tasks, such as room layout on cell-level search space.

\textbf{Evaluation of evolutionary algorithms with transferred population initialization. }We reproduced REA in our benchmark, and the result is presented in Figure \ref{fig:rea}. We randomly initialize a population during the pre-train phase, then set the pre-train budget, i.e., the total number of architecture to search during the pre-train phase, to be 50. After training on the given budget on a source task, we take the top 10 architectures in the pre-train history as the initialization of the population on the target task and search for 50 epochs. Although it does not have significant boosts on specific tasks like CATCH does, it maintains a relatively stable performance improvement across all examined tasks, which results in the slight surpass from its train-from-scratch version when the curves are averaged.

\textbf{Comparison across Transfer NAS Algorithms}. The average performance of each algorithm is presented in Table \ref{Table:comparison}. REA-transfer is the top performer among all evaluated algorithms, finding networks on the 99.02 and 99.11 percentile on the cell-level and macro-level search spaces. The experiments highlight that: (1) Direct transfer of architectures performs significantly worse than random search; (2) Direct policy transfer works better than direct architecture transfer, whereas it often results in negative transfer; (3) It is possible to improve the policy transfer's robustness with added mechanisms, such as CATCH's encoder and evaluator; (4) Maintaining consistent performance across tasks and search spaces remains a challenge for NAS algorithms. 

\section{Discussions and Conclusion}

\textbf{Major challenges of transferable NAS research.} After working closely with the benchmark, we realize that (1) the top networks can be very different across tasks. Therefore, the transfer schemes should be able to respond quickly if the task nature has significantly changed. However, effectively detecting and responding to such changes can be difficult. (2) transfer learning methods usually do not assume prior knowledge about future tasks, but if the policy is specifically designed for NAS, it is possible to incorporate certain NAS features to speed up learning. The major challenges lie in effectively designing such a scheme to provide the most relevant information.

\textbf{Suggestions for future NAS research:} (1) It is important to study efficient NAS strategies that work beyond cell-level search space, as some network attributes, such as the macro skeleton, might have a larger impact on performance for some tasks. (2) When transferring policies and architectures, including some carefully designed mechanisms might help tweak the transferred components toward directions favorable by the target task. (3) Evolutionary methods are not typical strategies studied by the transfer learning community, but its performance on the benchmark suggests that it might worth further investigation. Similarly, there might be other strategies outside of the pool of conventional transfer methods that are promising for transferable NAS research.

In this paper, we present TransNAS-Bench-101, a benchmark for improving the transferability and generalizability of NAS algorithms. We evaluate 7,352 neural networks on 7 vision tasks, provide detailed analysis on the benchmark, then point out challenges and suggestions for future research. It is difficult for algorithms to robustly maintain its performance when the task nature has shifted, and experiments show that there is still large room for improvement in NAS methods' generalizability. With this work, we hope to make cross-task NAS research more accessible and encourage more exceptional algorithms that are both efficient and flexible on multiple tasks and search spaces to evolve. In the future, we will try to (1) enlarge our search spaces and (2) evaluate all networks with more seeds. We welcome researchers to test their algorithms' generalizability on TransNAS-Bench-101, and we are happy to include their results in future versions of our benchmark.

{\small
\bibliographystyle{ieee_fullname}
\bibliography{egbib}

\begin{thebibliography}{10}\itemsep=-1pt

\bibitem{Mindspore}
Mindspore. https://www.mindspore.cn/en.

\bibitem{bergstra2012random}
James Bergstra and Yoshua Bengio.
\newblock Random search for hyper-parameter optimization.
\newblock {\em The Journal of Machine Learning Research}, 13(1):281--305, 2012.

\bibitem{cai2020tiny}
Han Cai, Chuang Gan, Ligeng Zhu, and Song Han.
\newblock Tiny transfer learning: Towards memory-efficient on-device learning.
\newblock {\em arXiv preprint arXiv:2007.11622}, 2020.

\bibitem{cai2018proxylessnas}
Han Cai, Ligeng Zhu, and Song Han.
\newblock Proxylessnas: Direct neural architecture search on target task and
  hardware.
\newblock In {\em ICLR}, 2019.

\bibitem{chen2018searching}
Liang-Chieh Chen, Maxwell Collins, Yukun Zhu, George Papandreou, Barret Zoph,
  Florian Schroff, Hartwig Adam, and Jon Shlens.
\newblock Searching for efficient multi-scale architectures for dense image
  prediction.
\newblock In {\em Advances in neural information processing systems}, pages
  8699--8710, 2018.

\bibitem{chen2020catch}
Xin Chen, Yawen Duan, Zewei Chen, Hang Xu, Zihao Chen, Xiaodan Liang, Tong
  Zhang, and Zhenguo Li.
\newblock Catch: Context-based meta reinforcement learning for transferrable
  architecture search.
\newblock {\em arXiv preprint arXiv:2007.09380}, 2020.

\bibitem{deng2009imagenet}
Jia Deng, Wei Dong, Richard Socher, Li-Jia Li, Kai Li, and Li Fei-Fei.
\newblock Imagenet: A large-scale hierarchical image database.
\newblock In {\em 2009 IEEE conference on computer vision and pattern
  recognition}, pages 248--255. Ieee, 2009.

\bibitem{dong2019bench}
Xuanyi Dong and Yi Yang.
\newblock Nas-bench-201: Extending the scope of reproducible neural
  architecture search.
\newblock In {\em International Conference on Learning Representations}, 2019.

\bibitem{dong2019searching}
Xuanyi Dong and Yi Yang.
\newblock Searching for a robust neural architecture in four gpu hours.
\newblock In {\em Proceedings of the IEEE Conference on computer vision and
  pattern recognition}, pages 1761--1770, 2019.

\bibitem{guobreaking}
Yong Guo, Yaofo Chen, Yin Zheng, Peilin Zhao, Jian Chen, Junzhou Huang, and
  Mingkui Tan.
\newblock Breaking the curse of space explosion: Towards efficient nas with
  curriculum search.

\bibitem{he2016deep}
Kaiming He, Xiangyu Zhang, Shaoqing Ren, and Jian Sun.
\newblock Deep residual learning for image recognition.
\newblock In {\em Proceedings of the IEEE conference on computer vision and
  pattern recognition}, pages 770--778, 2016.

\bibitem{isola2017image}
Phillip Isola, Jun-Yan Zhu, Tinghui Zhou, and Alexei~A Efros.
\newblock Image-to-image translation with conditional adversarial networks.
\newblock In {\em Proceedings of the IEEE conference on computer vision and
  pattern recognition}, pages 1125--1134, 2017.

\bibitem{kingma2014adam}
Diederik~P Kingma and Jimmy Ba.
\newblock Adam: A method for stochastic optimization.
\newblock {\em arXiv preprint arXiv:1412.6980}, 2014.

\bibitem{Li2018}
Zeming Li, Chao Peng, Gang Yu, Xiangyu Zhang, Yangdong Deng, and Jian Sun.
\newblock Detnet: A backbone network for object detection.
\newblock In {\em ECCV}, 2018.

\bibitem{Lian2020Towards}
Dongze Lian, Yin Zheng, Yintao Xu, Yanxiong Lu, Leyu Lin, Peilin Zhao, Junzhou
  Huang, and Shenghua Gao.
\newblock Towards fast adaptation of neural architectures with meta learning.
\newblock In {\em International Conference on Learning Representations}, 2020.

\bibitem{liang2019computation}
Feng Liang, Chen Lin, Ronghao Guo, Ming Sun, Wei Wu, Junjie Yan, and Wanli
  Ouyang.
\newblock Computation reallocation for object detection.
\newblock In {\em {ICLR}}. OpenReview.net, 2020.

\bibitem{lin2014microsoft}
Tsung-Yi Lin, Michael Maire, Serge Belongie, James Hays, Pietro Perona, Deva
  Ramanan, Piotr Doll{\'a}r, and C~Lawrence Zitnick.
\newblock Microsoft coco: Common objects in context.
\newblock In {\em European conference on computer vision}, pages 740--755.
  Springer, 2014.

\bibitem{liu2020labels}
Chenxi Liu, Piotr Doll{\'a}r, Kaiming He, Ross Girshick, Alan Yuille, and
  Saining Xie.
\newblock Are labels necessary for neural architecture search?
\newblock {\em arXiv preprint arXiv:2003.12056}, 2020.

\bibitem{liu2018darts}
Hanxiao Liu, Karen Simonyan, and Yiming Yang.
\newblock Darts: Differentiable architecture search.
\newblock In {\em International Conference on Learning Representations}, 2018.

\bibitem{mirza2014conditional}
Mehdi Mirza and Simon Osindero.
\newblock Conditional generative adversarial nets.
\newblock {\em arXiv preprint arXiv:1411.1784}, 2014.

\bibitem{miyato2018spectral}
Takeru Miyato, Toshiki Kataoka, Masanori Koyama, and Yuichi Yoshida.
\newblock Spectral normalization for generative adversarial networks.
\newblock {\em arXiv preprint arXiv:1802.05957}, 2018.

\bibitem{noroozi2016unsupervised}
Mehdi Noroozi and Paolo Favaro.
\newblock Unsupervised learning of visual representations by solving jigsaw
  puzzles.
\newblock In {\em European Conference on Computer Vision}, pages 69--84.
  Springer, 2016.

\bibitem{pasunuru2019continual}
Ramakanth Pasunuru and Mohit Bansal.
\newblock Continual and multi-task architecture search.
\newblock {\em arXiv preprint arXiv:1906.05226}, 2019.

\bibitem{real2019regularized}
Esteban Real, Alok Aggarwal, Yanping Huang, and Quoc~V Le.
\newblock Regularized evolution for image classifier architecture search.
\newblock In {\em Proceedings of the aaai conference on artificial
  intelligence}, volume~33, pages 4780--4789, 2019.

\bibitem{schulman2017proximal}
John Schulman, Filip Wolski, Prafulla Dhariwal, Alec Radford, and Oleg Klimov.
\newblock Proximal policy optimization algorithms.
\newblock {\em arXiv preprint arXiv:1707.06347}, 2017.

\bibitem{shaw2019meta}
Albert Shaw, Wei Wei, Weiyang Liu, Le Song, and Bo Dai.
\newblock Meta architecture search.
\newblock In {\em Advances in Neural Information Processing Systems}, pages
  11227--11237, 2019.

\bibitem{shi2020bridging}
Han Shi, Renjie Pi, Hang Xu, Zhenguo Li, James Kwok, and Tong Zhang.
\newblock Bridging the gap between sample-based and one-shot neural
  architecture search with bonas.
\newblock {\em Advances in Neural Information Processing Systems}, 33, 2020.

\bibitem{tan2019efficientnet}
Mingxing Tan and Quoc Le.
\newblock Efficientnet: Rethinking model scaling for convolutional neural
  networks.
\newblock In {\em International Conference on Machine Learning}, pages
  6105--6114, 2019.

\bibitem{wang2004image}
Zhou Wang, Alan~C Bovik, Hamid~R Sheikh, and Eero~P Simoncelli.
\newblock Image quality assessment: from error visibility to structural
  similarity.
\newblock {\em IEEE transactions on image processing}, 13(4):600--612, 2004.

\bibitem{wong2018transfer}
Catherine Wong, Neil Houlsby, Yifeng Lu, and Andrea Gesmundo.
\newblock Transfer learning with neural automl.
\newblock In {\em Advances in Neural Information Processing Systems}, pages
  8356--8365, 2018.

\bibitem{xu2019auto}
Hang Xu, Lewei Yao, Wei Zhang, Xiaodan Liang, and Zhenguo Li.
\newblock Auto-fpn: Automatic network architecture adaptation for object
  detection beyond classification.
\newblock In {\em Proceedings of the IEEE International Conference on Computer
  Vision}, pages 6649--6658, 2019.

\bibitem{yang2019evaluation}
Antoine Yang, Pedro~M Esperan{\c{c}}a, and Fabio~M Carlucci.
\newblock Nas evaluation is frustratingly hard.
\newblock In {\em International Conference on Learning Representations}, 2019.

\bibitem{yao2019sm}
Lewei Yao, Hang Xu, Wei Zhang, Xiaodan Liang, and Zhenguo Li.
\newblock Sm-nas: Structural-to-modular neural architecture search for object
  detection.
\newblock {\em arXiv preprint arXiv:1911.09929}, 2019.

\bibitem{ying2019bench}
Chris Ying, Aaron Klein, Eric Christiansen, Esteban Real, Kevin Murphy, and
  Frank Hutter.
\newblock Nas-bench-101: Towards reproducible neural architecture search.
\newblock In {\em International Conference on Machine Learning}, pages
  7105--7114, 2019.

\bibitem{zamir2018taskonomy}
Amir~R Zamir, Alexander Sax, William Shen, Leonidas~J Guibas, Jitendra Malik,
  and Silvio Savarese.
\newblock Taskonomy: Disentangling task transfer learning.
\newblock In {\em CVPR}, pages 3712--3722, 2018.

\bibitem{zhou2017places}
Bolei Zhou, Agata Lapedriza, Aditya Khosla, Aude Oliva, and Antonio Torralba.
\newblock Places: A 10 million image database for scene recognition.
\newblock {\em IEEE Transactions on Pattern Analysis and Machine Intelligence},
  2017.

\bibitem{zoph2018learning}
Barret Zoph, Vijay Vasudevan, Jonathon Shlens, and Quoc~V Le.
\newblock Learning transferable architectures for scalable image recognition.
\newblock In {\em Proceedings of the IEEE conference on computer vision and
  pattern recognition}, pages 8697--8710, 2018.

\end{thebibliography}
}

\clearpage



\twocolumn[{
    \centering
    {\Large \bf Supplementary Materials for TransNAS-Bench-101: Improving Transferability and Generalizability of Cross-Task Neural Architecture Search \par}
    \vspace*{24pt}
  }]

\setcounter{section}{0}

\section{Detailed Information of TransNAS-Bench-101 Benchmark Dataset}

We provide the train/validation/test performance information of each network at each epoch. One can also find each network's inference time, FLOPs, the total number of parameters, and time elapsed during each training epoch from the dataset. Each network's inference time is measured on one Tesla V100 with one image of shape (3, 720, 1080). FLOPs are computed with one image of shape (3, 224, 224).

\section{Training Details of Each Task} 
\textbf{Object Classification. }The labels provided by the Taskonomy dataset \cite{zamir2018taskonomy} are activations generated by a ResNet-152 model \cite{he2016deep} pre-trained on ImageNet \cite{deng2009imagenet}. For object classification, we train networks with the provided activations. Since we use a subset of the Taskonomy dataset, we identified 75 classes of objects that appear in our selected subset for network training. The data augmentations applied for this task are random flip, color jittering, and normalization. For each network, the decoder part contains a Global Average Pooling (GAP) layer and a linear layer. Referring to the settings of Taskonomy, each network is trained for 25 epochs. Throughout the learning process, we use a cosine annealing scheduler to gradually reduce the learning rate from 0.1 to 0 for fast convergence. The optimizer for parameters is SGD with the momentum factor 0.9, 0.0005 weight decay, and Nesterov momentum is enabled.

\textbf{Scene Classification. }Similar to Object Classification, the Taskonomy dataset's labels for scene classification comes from an ImageNet pre-trained ResNet-152 model. Our selected dataset contains 47 classes out of the original 365 classes. Referring to the settings of Taskonomy, each network is trained for 25 epochs. The data augmentation, decoder, optimizer, and learning rate scheduler settings are the same as Object Classification tasks.

\textbf{Room Layout. }The goal of this task is to estimate and align a 3D bounding box. In the Taskonomy dataset, such a bounding box is defined by a 9-dimension vector. The network is updated through computing the Mean Square Error (MSE) loss with the provided labels. The data augmentation methods used are color jittering and normalization. Following the settings of Taskonomy, each network is trained for 25 epochs. The decoder, optimizer, and learning rate scheduler settings are the same as Object Classification and Scene Classification.

\textbf{Jigsaw Content Prediction. } Jigsaw's inclusion is inspired by a recent work \cite{liu2020labels} that explores the potential of self-supervised tasks in architecture search. We follow \cite{noroozi2016unsupervised} to design the self-supervised task Jigsaw. The input image is divided into 9 patches and shuffled according to one of 1000 preset permutations. The goal of this task is to classify which permutation is used. We use a Siamese network to extract the feature map of each of the 9 image tiles and concatenate them. We apply random flip, color jitter, and random grayscale with a probability of 0.3 for data augmentation. Referring to the settings of Taskonomy, each network is trained for 10 epochs since Jigsaw tasks converge very quickly. The decoder, optimizer, and learning rate scheduler settings are the same as above.

\textbf{Semantic Segmentation. }The labels provided by the Taskonomy dataset on semantic segmentation are generated through a network pre-trained on the MSCOCO \cite{lin2014microsoft} dataset. Our selected subset contains 17 semantic classes. We apply random flip, color jitter, and normalization for data augmentation. For this task, we use the SGD optimizer with a learning rate 0.1, along with a cosine annealing scheduler. Similar to the settings in Taskonomy, each network is trained for 30 epochs. The decoder, optimizer, and learning rate scheduler settings are the same as above.

\begin{figure*}[!b]
    \centering
    \subfigure[Cell-level]
    {
        \includegraphics[height=0.95\columnwidth]{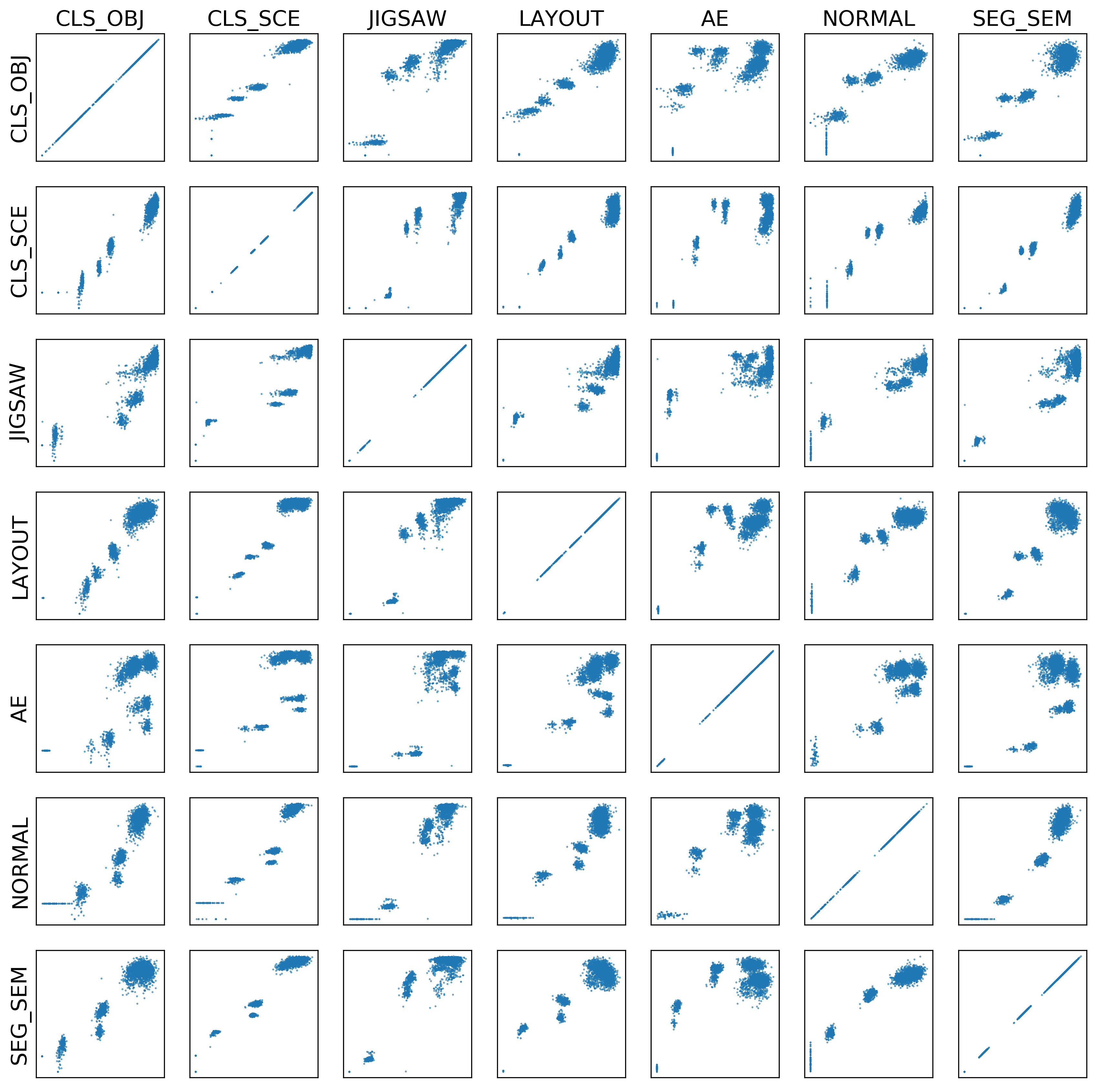}
        \label{fig:all-1}
    }
    \hspace{8mm}
    \subfigure[Macro-level]
    {
        \includegraphics[height=0.95\columnwidth]{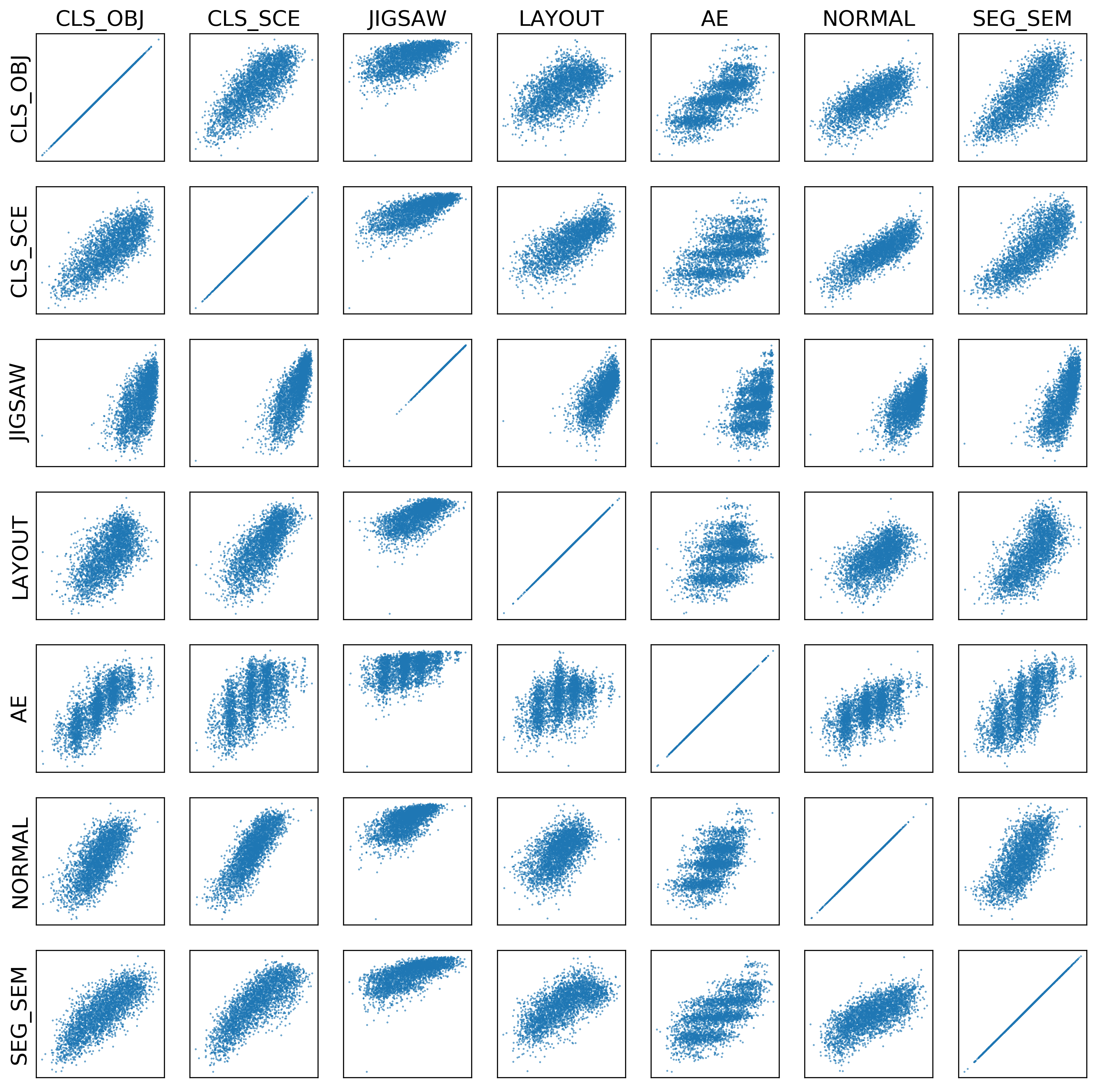}
        \label{fig:all-2}
    }
    \caption{Performance of cell-level and macro-level networks across all seven tasks.}
    \label{fig:ranking}
\end{figure*}

\textbf{Autoencoding. }The generator networks in the Autoencoding task follow an encoder-decoder structure in Pix2Pix \cite{isola2017image}, where the encoders are the searched backbones and the decoders contain 14 layers of convolution and deconvolution. We train the generator network using conditional GAN \cite{mirza2014conditional} with a discriminator containing 7 convolution layers. We apply spectral normalization \cite{miyato2018spectral} to stabilize the discriminator. The generator is trained with the L1 loss with weight 0.99 and GAN loss with weight 0.01. We use structural similarity index measure (SSIM) \cite{wang2004image} as the metric for network performance evaluation. The data augmentations applied for the generator are random flip and color jittering.  Both the generator and discriminator use Adam \cite{kingma2014adam} optimizer to stabilize training with an initial learning rate of 0.0005. Similar to the settings in Taskonomy, each network is trained for 30 epochs.

\textbf{Surface Normal. }We use the same generator, discriminator, evaluation metric, and loss for surface normal and autoencoding tasks. For surface normal, the optimizers for both generator and discriminator are Adam with a learning rate of 0.0001. Referring to the settings in Taskonomy, each network is trained for 30 epochs.

\section{Cross-task Training Results}
We plot the network performance relations for all tasks in Figure \ref{fig:ranking}. Networks in the cell-level search space have a much greater performance gap than networks in the macro-level search space because certain cell designs can easily lead to poor network performance (e.g., choosing skip-connection for all operations). The network performance in most tasks is positively correlated. 

\section{Convergence Analysis of Trained Networks}

\begin{figure*}[t]
\vspace{-3mm}
     \begin{center}
     \subfigure[Cell-level search space]
     {
        \includegraphics[width=2\columnwidth]{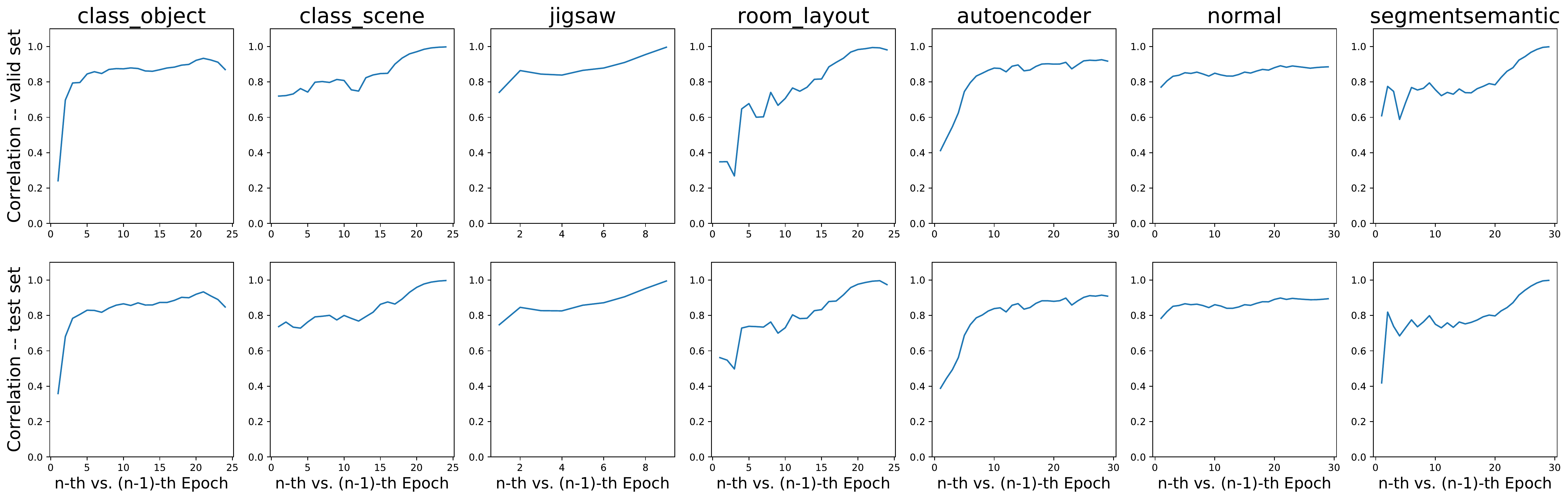}
        \label{fig:Cell-level-converge}
     }
     \subfigure[Macro-level search space]
     {
        \includegraphics[width=2\columnwidth]{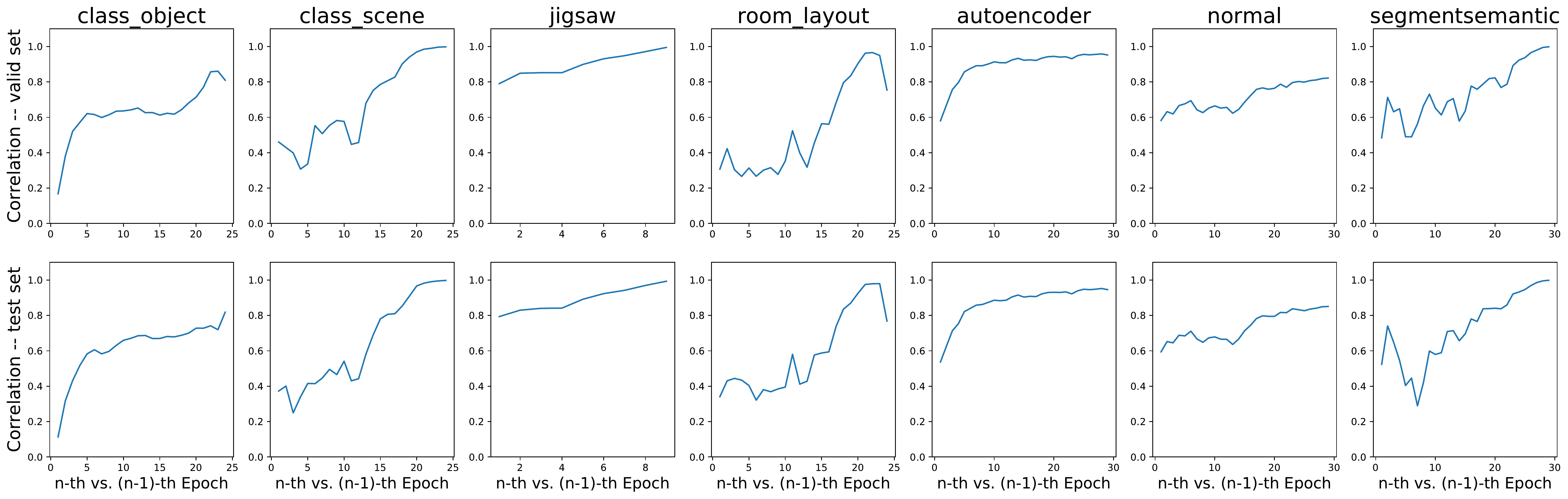}
        \label{fig:Macro-level-converge}
     }
    \caption{Figure \ref{fig:Cell-level-converge}-\ref{fig:Macro-level-converge} show the network rank correlation between the $n$-th epoch and ($n-1$)-th epoch on the cell-level search space and the macro-level search space. The first row in both subfigures are correlations on the validation set, and the second rows are correlations on the test set.}
    \label{fig:converge}
    \end{center}
    \vspace{-3mm}
\end{figure*}

To show the extent to which the ranking of networks in our search space has stabilized, we plot the network ranking correlation between consecutive epochs in Figure \ref{fig:converge}. We query the network performance at each epoch, then calculate the network ranking correlation between epoch $t$ and epoch $t-1$. A higher value at epoch $t$ means that this additional training epoch does not significantly change the relative advantage of each network in the search space. We plot such correlation on all tasks in Figure \ref{fig:converge}. From the figures, we can see that the network rankings in most tasks tend to stabilize as they approach the end of the training. Despite the relatively short training budget for each task, the networks have displayed good convergence results.

\section{Algorithm Training Details}

We implemented four baseline algorithms using TransNAS-Bench-101 and provide implementation details of each algorithm below.

\textbf{Regularized Evolution for Image Classifier Architecture Search (REA)}. \cite{real2019regularized}. For both search spaces, the population size and sample size are set to 10. We set the number of cycles as 40. Hence, a total of 50 architectures would be selected. The best validation accuracy throughout the evaluation of a network would be chosen as the fitness. For the macro-level search space, the mutation operation would be randomly adding, deleting, or changing a module type.

\textbf{Proximal Policy Optimization (PPO)}. \cite{schulman2017proximal}. For both search spaces, we can formulate the NAS problem as a sequential decision problem, and the reinforcement learning algorithm PPO aims to select each attribute choice to form a network. We set the learning rate as 0.01, and it decays by 0.999 for every 15 steps. The optimizer is Adam. We set the clipping parameter $\epsilon$ as 0.2, memory size as 100, discount $\gamma$ as 0.99, GAE parameter $\lambda$ as 0.95, value function coefficient as 1, and entropy coefficient as 0.01. For PPO-transfer, we first pre-train the policy by applying the policy to search on the lowest cost tasks, jigsaw, then transfer to target tasks.

\textbf{Context-based Meta Reinforcement Learning for Transferrable Architecture Search (CATCH)}. \cite{chen2020catch}. CATCH uses PPO as its controller to sample networks. It also uses a network evaluator to predict the network performance and uses a context encoder to learn a task-specific embedding to guide the search. CATCH incorporated meta reinforcement learning by first meta-training the policy on various tasks, such as jigsaw, classification tasks, and then adapt the meta-trained policy to a target task. We set the hyperparameters for the controller the same as those of PPO. For the context encoder, we set its learning rate as 0.0005, KL Divergence weight as 0.1. For the evaluator, we set its learning rate as 0.0005, initial epsilon $\epsilon$ as 1, and it decays by 0.025 for every 4 steps to encourage exploration. In the adaptation phase, the initial epsilon for the evaluator is set to 0.5 and decays by 0.025 for every 2 steps to encourage exploitation.

\end{document}